%% file: SBP.tex
\documentclass[10pt,journal,compsoc]{IEEEtran}
\usepackage{ifthen}

\ifCLASSOPTIONcompsoc
  \usepackage[nocompress]{cite}
\else
  \usepackage{cite}
\fi
\ifCLASSINFOpdf
\else
\fi

\usepackage{amsmath,amssymb,amsthm}
\usepackage{bm}
\usepackage[norelsize,ruled,vlined]{algorithm2e}

\ifCLASSOPTIONcompsoc
 \usepackage[caption=false,font=footnotesize,labelfont=sf,textfont=sf]{subfig}
\else
 \usepackage[caption=false,font=footnotesize]{subfig}
\fi

\usepackage{float}

\usepackage{color}
\usepackage{pgfplots}

\usepackage{times}
\usepackage{multirow}
\usepackage{booktabs}

\newcommand{\vm}[1]{\ensuremath{\bm{#1}}}

\newcommand{\Real}{\mathbb{R}}
\newcommand{\RealPositive}{\mathbb{R}_+^*}
\newcommand{\bigO}{\mathcal{O}}
\newcommand{\pseudomarginalsExact}[1][]{\ifthenelse{\equal{#1}{}}{{P_{B}}}{P_{\mathbf{X}_{B,[#1]}}}}
\newcommand{\pseudoTemplate}[2]
{\ifthenelse{\equal{#1}{}}
  {{\tilde{P}_{B}^{#2}}} {{\tilde{P}^{#2}_{{B},#1}}}}
\newcommand{\pseudomarginals}[1][]			
  {\ifthenelse{\equal{#1}{}}
  {\pseudoTemplate{}{}} {\pseudoTemplate{#1}{}}}
\newcommand{\pseudomarginalsMinLocal}[1][]
  {\ifthenelse{\equal{#1}{}}
  {\pseudoTemplate{}{\circ}}{\pseudoTemplate{#1}{\circ}}}

\newcommand{\pseudomarginalsMinLocalNeg}[1][]{\ifthenelse{\equal{#1}{}}{{\tilde{P}_{\mathbf{X}_{\mathcal{B}}}}^{\oplus}}{{\tilde{P}_{\mathbf{X}_{\mathcal{B}[#1]}}}^{\oplus}}}
\newcommand{\pseudomarginalsMinGlobal}[1][]
{\ifthenelse{\equal{#1}{}}
  {\tilde{P}_B^*} {\tilde{P}_{B,#1}^*}}
\newcommand{\marginals}[1]{P_{#1}}
\newcommand{\marginalsApprox}[1]{{\tilde{P}_{#1}}}

\newcommand{\marginalsShort}[1]{P(#1)}
\newcommand{\marginalsShortApprox}[1]{\tilde{P}(#1)}
\newcommand{\stateSpace}{\mathcal{X}}

\DeclareMathOperator{\E}{\mathbb{E}}
\DeclareMathOperator*{\argmin}{argmin}
\newcommand{\graph}{\mathcal{G}}
\newcommand{\setOfNodes}{\mathbf{X}}

\newcommand{\setOfEdges}{\mathbf{E}}
\newcommand{\RV}[1]{X_{#1}}

\newcommand{\edge}[2]{(#1,#2)}			
\newcommand{\neighbors}[1]{N({#1})}
\newcommand{\PGM}[1][]{\mathcal{U}_{#1}}
\newcommand{\pairwisePot}[3][]{\ifthenelse{\equal{#1}{}}{{\Phi_{X_{#2},X_{#3}}(x_{#2},x_{#3})}} {\Phi_{X_{#2},X_{#3}}(x_{#2},x_{#3})_{[#1]} }}
\newcommand{\pairwiseShort}[2]{\Phi(x_{#1},x_{#2})}
\newcommand{\pairwiseSBP}[3]{\Phi_{#3}(x_{#1},x_{#2})}
\newcommand{\localPot}[2][]{\ifthenelse{\equal{#1}{}} {\Phi_{X_{#2}}(x_{#2})} {\Phi_{X_{#2}}(x_{#2})_{[#1]}}}
\newcommand{\localShort}[1]{\Phi(x_{#1})}
\newcommand{\localSBP}[2]{\Phi_{#2}(x_{#1})}

\newcommand{\setOfPot}{ \Psi}
\newcommand{\coupling}[2]{J_{#1#2}}
\newcommand{\field}[1]{\theta_{#1}}
\newcommand{\mean}[1]{m_{#1}}
\newcommand{\meanMinLocal}[1]{m_{#1}^{\circ}}

\newcommand{\meanMinLocalSpecific}[2]{m_{#1}^{#2}}
\newcommand{\correlation}[2]{\chi_{#1#2}}
\newcommand{\correlationMinLocalSpecific}[3]{\chi_{#1#2}^{#3}}

\newcommand{\BP}{\mathcal{BP}}
\newcommand{\BPD}{\text{BP}_{\text{D}}}
\newcommand{\msg}[4][]{\ifthenelse{\equal{#4}{}} {\mu^{#1}_{#2 #3}(x_{#3})} {\mu^{#1}_{#2 #3}(\RV{#3}=#4)}}
\newcommand{\msgShort}[4][]{\ifthenelse{\equal{#4}{}} {\mu^{#1}_{#2 #3}(x_{#3})} {\mu^{#1}_{#2 #3}(#4)}}
\newcommand{\setOfMsg}[1][]{\bm{\mu}^{#1}}

\newcommand{\msgRatio}[3][]{\bar{\mu}_{#2 #3}}
\newcommand{\scaling}{\zeta}
\newcommand{\SBP}{\text{SBP}}
\newcommand{\startp}{\pseudomarginalsMinLocal(\scaling=0)}
\newcommand{\terminalp}{\pseudomarginalsMinLocal(\scaling=1)}

\newcommand{\mse}[1][]{\text{MSE}}
\newcommand{\mseb}{\text{MSE}_{{B}}}
\newcommand{\FGibbs}{\mathcal{F}}
\newcommand{\FB}[1][]
  {\ifthenelse{\equal{#1}{}}
  {\mathcal{F}_B} {\mathcal{F}_{B,all}}}
\newcommand{\FBMinGlobal}{\mathcal{F}_B^*}
\newcommand{\FBMinLocal}{\mathcal{F}_B^{\circ}}

\newcommand{\FBMinLocalSpecific}[1]{\mathcal{F}_B^{#1}}
\newcommand{\EB}{E_{{B}}}
\newcommand{\SB}{S_{{B}}}

\newcommand{\PartitionGibbs}{\mathcal{Z}}

\newcommand{\PartitionApprox}{\mathcal{{Z}_B}}
\newcommand{\LP}[1][]{\mathcal{L_{#1}}}

\newcommand{\init}{\text{init}} 

\newtheorem{thm}{Theorem}
\newtheorem{lm}[thm]{Lemma}
\newtheorem{con}{Conjecture}
\newtheorem{cor}{Corollary}[thm]
\newtheorem{prop}{Properties}

\theoremstyle{definition}
\newtheorem{ex}{Example}

\newcommand{\iteration}[1]{{#1}}

\newcommand{\stableOne}{s}
\newcommand{\stableTwo}{t}

\begin{document}
\title{Self-Guided Belief Propagation -- A Homotopy Continuation Method}
%
%
%
%

\author{Christian~Knoll,~\IEEEmembership{Associate~Member,~IEEE,}
        Adrian~Weller,
        and~Franz~Pernkopf,~\IEEEmembership{Senior~Member,~IEEE}
        
\IEEEcompsocitemizethanks{\IEEEcompsocthanksitem Christian Knoll (christian.knoll.c@ieee.org), and Franz Pernkopf (pernkopf@tugraz.at) are with the Signal Processing and Speech Communication Laboratory, Graz University of Technology.
\IEEEcompsocthanksitem Adrian Weller (aw665@cam.ac.uk) is with the University of Cambridge and the Alan Turing Institute.\protect\\
}
}

\markboth{Self-Guided Belief Propagation -- A Homotopy Continuation Method}%
{}
\IEEEtitleabstractindextext{%
\begin{abstract}
Belief propagation (BP) is a popular method for performing probabilistic inference on graphical models.
In this work, we enhance BP and propose self-guided belief propagation (SBP) that incorporates the pairwise potentials only gradually.
This homotopy continuation method converges to a unique solution and increases the accuracy without increasing the computational burden. 
We provide a formal analysis to demonstrate that SBP finds the global optimum of the Bethe approximation for attractive models where all variables favor the same state.
Moreover, we apply SBP to various graphs with random potentials and empirically show that: 
(i) SBP is superior in terms of accuracy whenever BP converges, and 
(ii) SBP obtains a unique, stable, and accurate solution whenever BP does not converge. 
\end{abstract}

\begin{IEEEkeywords}
Graphical models, belief propagation, probabilistic inference, sum-product algorithm, partition function, inference algorithms.
\end{IEEEkeywords}}

\maketitle

\IEEEdisplaynontitleabstractindextext
\IEEEpeerreviewmaketitle

\IEEEraisesectionheading{\section{Introduction}\label{sec:introduction}}

\IEEEPARstart{C}{omputing} the marginal distributions and evaluating the partition function are two fundamental problems in the context of probabilistic graphical models.
Both problems can be solved efficiently for tree-structured models but are NP-hard if the graphical model contains loops~\cite{cooper1990computational}.
Since loops are present in many problems of relevance there is a need for efficient approximation methods.

Belief propagation (BP) provides a way to approximate the marginal distribution and may be viewed as an approach to try to minimize the Bethe free energy $\FB$~\cite{yedidia}.
BP has a long success story in many applications, including computer vision, speech processing, social network analysis, and error-correcting codes~\cite{koller-friedman,jordan2004,pernkopf2014pgm}. 
But, despite its success, BP does not always approximate the marginals well and may even fail to converge.
Reasons for BP's failure include the existence of 
multiple fixed points of varying accuracy (where it depends on implementation details to which one BP converges) and of  unstable fixed points that cause BP to oscillate far away from any fixed point~\cite{weiss2000correctness, mooij2007sufficient, ihler2005loopy}.
These limitations motivate the search for modifications of BP that have better convergence properties and find more accurate marginals.

One alternative is to perform a different optimization problem that minimizes the Bethe free energy $\FB$, instead of applying BP~\cite{yedidia}.
While, for certain problem classes, polynomial-time algorithms exist, it is even problematic to approximate the global minimum  for models with arbitrary potentials~\cite{chandrasekaran2011counting, shin2012complexity, weller2013approximating}. 
Hence, the pursuit for methods that approximate the marginals well -- ideally with both run-time and convergence guarantees -- is still ongoing.

In this work, we present self-guided belief propagation (SBP) to help address this gap. 
The observation that strong pairwise potentials often reduce accuracy and worsen the convergence properties~\cite{knoll2017uai} inspired us to 
construct a homotopy; i.e., we first consider only local potentials (where BP is exact) and subsequently modify the model by gradually increasing the pairwise potentials to their given values.
SBP thus solves a deterministic sequence of models that iteratively refines the Bethe approximation towards a solution that is uniquely defined by the initial model.

We empirically evaluate SBP for grid-graphs, complete graphs, and random graphs with random potentials.
Compared to BP, we observe superior performance in terms of accuracy; in fact, SBP achieves more accurate results than Gibbs sampling in a fraction of run-time.
Additionally, SBP exhibits favorable convergence properties and excels for hard problems, for which -- despite BP failing to converge -- SBP provides accurate results.
We also analyze SBP theoretically and demonstrate optimality of the selected fixed point for \emph{attractive} models with unidirectional local potentials.\footnote{An attractive model is a probabilistic graphical model where all pairwise interactions are specified by positive couplings; a general model also contains pairwise interactions with negative couplings. Unidirectional local potentials are potentials that all favor the same state (see Section~\ref{subsec:background:model}).}

It is worth emphasizing that SBP is a relatively straightforward modification to standard BP that can be easily added to existing BP implementations.
We thus expect that the ease of use lowers the hurdle for practical applications.

The paper is structured as follows: Section~\ref{sec:background} provides background on probabilistic graphical models, belief propagation, and methods that minimize the free energy. 
Moreover, we present a unified taxonomy of model-classes according to their complexity.
Our proposed algorithm is presented in Section~\ref{sec:SBP}. 
We evaluate SBP and discuss empirical observations in Section~\ref{sec:experiments} and provide a more formal analysis in Section~\ref{sec:theory}. 
Finally, we conclude the paper in Section~\ref{sec:conclusion}.


\section{Background}\label{sec:background}
In this section, we briefly introduce probabilistic graphical models and specify the models considered in this work. 
We further introduce the BP algorithm and its connection to the Bethe approximation. 
Finally, we summarize important properties regarding the solution space of BP as a foundation for our theoretical analysis of SBP.

\subsection{Probabilistic Graphical Models}
Let us  consider an undirected graph $\graph = (\setOfNodes,  \setOfEdges)$, where $\setOfNodes = \{\RV{1},\dots,\RV{N}\}$ is the set of nodes, and $\setOfEdges$ is the set of undirected edges. 
Then, two nodes $\RV{i}$ and $\RV{j}$ are joined by an edge if $\edge{i}{j}\in \setOfEdges$; note that we consider each edge only once, i.e., $\edge{i}{j} = \edge{j}{i}$. 
We further denote the set of neighbors of $\RV{i}$ by $\neighbors{i} = \{\RV{j} \in \setOfNodes : \edge{i}{j} \in \setOfEdges \}$.

Let us define a probabilistic graphical model $\PGM = (\graph,\setOfPot)$, where $\setOfPot = \{\Phi({\mathbf{x}_1}), \ldots, \Phi({\mathbf{x}_K})\}$ is the set of all $K$ potentials and $\setOfNodes$ is the set of random variables.
In this work we focus on pairwise models where all potentials consist of two variables at most, so that the joint distribution $\marginals{\setOfNodes}(\mathbf{x})$ factorizes according to 
\begin{align}
 \marginals{\setOfNodes}(\mathbf{x})  &=  \frac{1}{\PartitionGibbs} \prod_{\edge{i}{j} \in \setOfEdges} \pairwiseShort{i}{j}\prod_{i=1}^N \localShort{i}\label{eq:jointApprox} \\
 &= \frac{1}{\PartitionGibbs} \exp\big(-E(\mathbf{x})\big) \nonumber,
\end{align}
where the potentials specify the energy $E(\mathbf{x})$.

We consider the following two problems: 
First, evaluating the partition function $\PartitionGibbs$ (i.e., the normalization function of the joint distribution).
Consider any distribution $Q_{\mathbf{X}}(\mathbf{x})$ and the (Gibbs) free energy
$\FGibbs= \FGibbs(Q_{\mathbf{X}}(\mathbf{x}))=\sum_{\mathbf{x}} Q_{\mathbf{X}}(\mathbf{x}) \big(E(\mathbf{x}) + \ln Q_{\mathbf{X}}(\mathbf{x})\big)$;
then, evaluating the partition function and minimizing the free energy are equivalent since the minimum of $\FGibbs$ corresponds to $\PartitionGibbs$ according to $\min \FGibbs = \FGibbs^* = -\ln \PartitionGibbs$ (cf.~\cite{yedidia}).

Second, we consider computing the marginal distribution
\begin{align}
\marginals{\mathbf{Y}}(\mathbf{y}) = \sum_{x_i : \RV{i} \in \{\setOfNodes \backslash \mathbf{Y}\}} \marginals{\setOfNodes}(\mathbf{x}),
\end{align}
where $\mathbf{Y} \subset \setOfNodes$ may be any set of RVs.
The identities of the random variables are often obvious from the notation of the values; 
therefore, we often use the shorthand notation $\marginalsShort{\mathbf{y}}$ for the marginal distribution. 
Note that the problems of evaluating the partition function and computing the marginal distributions are inherently linked, as $\FGibbs$ obtains its minimum precisely for the marginal distributions $\marginals{\setOfNodes}(\mathbf{x})$.

In general, both problems are intractable~\cite{cooper1990computational}.
Consequently one often relaxes the problems and only approximates the marginal distributions and the partition function.
This allows for an elegant iterative algorithm that was discovered multiple times in different fields. 
It is known as belief propagation (BP) in computer science, the sum-product algorithm in information theory, and the cavity- or the Bethe-method in physics; we refer the reader to~\cite{kschischang2001factor,mezard2009} for a good overview of the underlying principles.

\subsection{MODEL SPECIFICATION}\label{subsec:background:model}
We consider binary pairwise models where every random variable $\RV{i}$ takes values from $\stateSpace = \{-1,+1\}$ (i.e., Ising models).
For these models, instead of considering the singleton marginals $\marginalsShortApprox{x_i}$ and the pairwise marginals $\marginalsShortApprox{x_i,x_j}$ explicitly, it is often more convenient to work with the means $\mean{i}$ and the correlations $\correlation{i}{j}$ defined according to
\begin{align}
 \mean{i} = \E(\RV{i}) = \marginals{\RV{i}}(+1) - \marginals{\RV{i}}(-1),
\end{align}
\begin{align}
\correlation{i}{j} = \E(\RV{i}, \RV{j}) = \sum_{x_i,x_j}x_i x_j \marginalsShort{x_i,x_j}.
\end{align}
Let us define couplings $\coupling{i}{j} \in \Real$ 
that are
assigned to each edge $\edge{i}{j} \in \setOfEdges$ and local fields $\field{i} \in \Real$ that act on each variable $\RV{i} \in \setOfNodes$. 
These parameters then define the pairwise potentials $\pairwiseShort{i}{j} = \exp(\coupling{i}{j}x_i x_j)$ and the local potentials $\localShort{i} = \exp(\field{i} x_i)$ so that the corresponding joint distribution from \eqref{eq:jointApprox} can be written as
\begin{align}
  \marginals{\setOfNodes}(\mathbf{x})  &= \frac{1}{\PartitionGibbs} \exp \left( \sum_{\edge{i}{j} \in \setOfEdges} J_{ij} x_i x_j + \sum_{i=1}^N \theta_i x_i \right).
\label{eq:joint}
\end{align}

There are two different types of interactions between random variables: 
if $\coupling{i}{j}$ is negative then the edge $\edge{i}{j}$ is \emph{repulsive}; 
if $\coupling{i}{j}$ is positive then the edge $\edge{i}{j}$ is \emph{attractive}. 
Accordingly, one distinguishes between \emph{general} models that contain both attractive and repulsive edges and \emph{attractive} models\footnote{
Note that attractive models are also known as ferromagnetic models~\cite{mezard2009} or log-supermodular models~\cite{ruozzi2013bethebound}.} 
that contain only attractive edges.

Attractive models are often more well-behaved than general models but, rather than only distinguishing between these two model-types, we find it beneficial to consider an even finer-grained distinction into the following three model-classes.
In doing so, we additionally unify different naming conventions in the literature and provide a consistent taxonomy of model-classes. 

\subsubsection{Frustrated, Balanced, and Unidirectional Models}\label{subsec:model:classes}
\begin{figure}[!t]
\centering
	\subfloat[][]{\label{fig:modelClasses1} \includegraphics[width=0.20\linewidth]{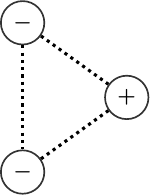}}
	\hspace{0.02\linewidth}
	\subfloat[][]{\label{fig:modelClasses2} \includegraphics[width=0.20\linewidth]{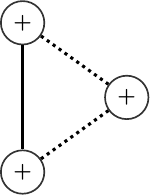}}
	\hspace{0.02\linewidth}
		\subfloat[][]{\label{fig:modelClasses2_2} \includegraphics[width=0.20\linewidth]{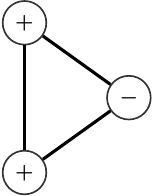}}
	\hspace{0.02\linewidth}
	\subfloat[][]{\label{fig:modelClasses3} \includegraphics[width=0.20\linewidth]{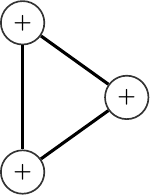}}
\caption{Illustration of the model-classes specified in Section~\ref{subsec:model:classes}:
(a) frustrated; (b) balanced, and (c) its equivalent attractive model; and (d) unidirectional model.
Solid lines depict attractive edges and dashed lines depict repulsive ones.
The signs in the vertices are equal to the signs of the corresponding local fields $\field{i}$. }
\end{figure}
First, \emph{frustrated models} are models that contain cycles for which the product over all couplings equates to a negative number, i.e., cycles with an odd number of repulsive edges (see Fig.~\ref{fig:modelClasses1}).\footnote{ 
Frustrated models are also referred to as spin glasses in the physics literature. These are some of the most complex models considered of the form~\eqref{eq:joint}.}
Note that if a model is frustrated, it must be a general model per definition.

Second, \emph{balanced models} are models that are not frustrated.
For a balanced model, it is possible to flip a certain subset of variables without affecting the represented distribution such that the resulting model is attractive (cf.~\cite{weller2015balanced}).
To flip a variable $\RV{i}$, one must switch its two states, and: reverse the sign of the corresponding local field $\field{i}$;  
and reverse the signs of the couplings $\coupling{i}{j}$ for all incident edges $\edge{i}{j}:\RV{j}\in \neighbors{i}$.
For an illustrative example, consider the model in Fig.~\ref{fig:modelClasses2}.
This model is a general one in its current form; 
reversing the sign of the rightmost variable further implies changing both attached edges from repulsive to attractive, which results in an equivalent attractive model (see Fig.~\ref{fig:modelClasses2_2}).
Such an equivalent model exists if and only if the original model contains no cycles with an odd number of repulsive edges, i.e., if the original model was not frustrated~\cite{harary1953}.

Finally, \emph{unidirectional models} are attractive models with unidirectional local potentials, i.e., either all $\field{i} \leq 0$ or $\field{i} \geq 0$ (see Fig.~\ref{fig:modelClasses3}).\footnote{ Unidirectional models are also referred to as attractive models in an external field, true ferromagnetic models, or consistent models in the literature.}
In a similar manner as for balanced models, certain general models have equivalent unidirectional models, i.e., by flipping variables to render the model attractive all local potentials are mirrored so that the attractive model is unidirectional; we call these models \emph{quasi-unidirectional}. 
All theoretical findings for unidirectional models thus readily extend to quasi-unidirectional models.\\

The major advantage of considering unidirectional models is that they are relatively well understood.
Thus unidirectional models are of particular relevance for our work -- as well as for studying inference algorithms in general.

Intuitively, 
in unidirectional models, we expect that all variables will energetically favor the same state as indicated by the local potentials.
Additionally, since all edges are attractive, no contradictions between neighboring variables will occur, thus further reinforcing this behavior.
Moreover, various research directions indicate the relevance of this model class. We will list a couple of these indications below:

On the one hand, it is generally often easier to study homogeneous models, where all pairwise potentials are the same (i.e., $\coupling{i}{j} = \coupling{}{}$) and all local potentials are the same (i.e., $\field{i} = \field{}$). 
Provided that the model is unidirectional, the findings from homogeneous models carry over readily~\cite{knoll2017uai, goldberg2003computational}.
Alternatively, one often considers models of the form~\eqref{eq:joint} in an external field $\field{}$, i.e, where all local potentials are the same (or have the same sign)~\cite{jerrum1993polynomial}. 
Keeping the physical interpretation in mind, we might hope that models that are placed in an external field are easier to understand.
Let us use the concept of flipping variables for transforming balanced models into attractive models again.
Interestingly, if we consider models in an external field only, then flipping variables is not a viable option as this would imply reversing the sign of some $\field{i}$.
Thus for models in a constant external field, the class of attractive models reduces precisely to the class of unidirectional models.

On the other hand, the complexity of approximating the partition function also depends on the model class and is in line with our distinction. 
That is for unidirectional models the partition function can be approximated efficiently in polynomial time~\cite{jerrum1993polynomial}; 
for frustrated models, the partition function cannot be approximated efficiently; balanced models are of intermediate complexity ~\cite{goldberg2003computational}.

Finally, it is sometimes beneficial to study models without local potentials;
in fact, one can find an auxiliary model without local potentials for every model of the form~\eqref{eq:joint} by adding one variable to the graph (and connecting it to all existing variables), where the auxiliary model has the same marginals and the same partition function (up to a constant factor of two)~\cite{weller2016uprooting,johnson2016learning, saade2017spectral}.
Provided that the original model was attractive, then the auxiliary model remains attractive (or can be rendered attractive by flipping the additional variable) if and only if the original model was unidirectional.\footnote{ 
On the one hand, we need to allow for a flip of the additional variable for the case that all $\field{i}<0$; on the other hand, this makes the statement agnostic to the definition (i.e., it holds irrespective if we have $\stateSpace = \{-1,+1\}$ or $\stateSpace = \{0,1\}$).}

\subsection{BELIEF PROPAGATION (BP)}\label{subsec:background:bp}
BP approximates the marginals by recursively exchanging messages between random variables. 
The messages $\msg[n+1]{i}{j}{}$ from $\RV{i}$ to $\RV{j}$ at iteration $n+1$ are updated according to \eqref{eq:update}
and are normalized so that $\sum_{x_j \in \stateSpace} \msg[n]{i}{j}{} = 1$.
\begin{align}
\msg[n+1]{i}{j}{} \propto \sum \limits_{x_i \in \stateSpace}  \pairwiseShort{i}{j}\localShort{i} \!\!\! \prod \limits_{\RV{k} \in \neighbors{i} \backslash {j}} \!\!\! \msg[n]{k}{i}{}
 \label{eq:update}
\end{align}

Let $\setOfMsg[n] = \{\msg[n]{i}{j}{}  : e_{ij}  \in \mathbf{E}, x_j \in \stateSpace \}$ be the set of all messages at iteration $n$ and let the mapping induced by~\eqref{eq:update} be denoted as
$\setOfMsg[n+1] = \BP(\setOfMsg[n])$.
If all successive messages remain unchanged, i.e., if $\setOfMsg[n+1] = \setOfMsg[n]$,  then BP is converged to a \emph{fixed point} $\setOfMsg[\circ]$. We further write $\setOfMsg[\circ] = \BP^{\circ}(\setOfMsg[\init])$, where $ \BP^\circ$ updates the messages until convergence.
If BP fails to converge and the messages oscillate, one can try to achieve convergence by either changing the update-rule~\cite{elidan2012residual,sutton2012improved,knoll2015message}, or by replacing the messages with a convex combination of the last messages~\cite{murphy1999loopy}. The latter method is known as damping ($\BPD$) where a damping parameter $\epsilon \in [0,1)$ specifies the new update rule
\begin{align}
  \setOfMsg[n+1] &= \mathcal{BP}_D(\setOfMsg[n]),\nonumber \\
                 &= (1-\epsilon)\BP(\setOfMsg[n])+\epsilon \setOfMsg[n].
\end{align}
After convergence, the singleton marginals $\marginalsShortApprox{x_i}$ and pairwise marginals $\marginalsShortApprox{x_i,x_j}$ are approximated by 
\begin{align}
\marginalsShortApprox{x_i} =& \frac{1}{Z_i} \localShort{i} \prod_{\RV{k} \in \neighbors{i}} \msg[\circ]{k}{i}{}, \label{eq:marginalsSingle} \\
\marginalsShortApprox{x_i,x_j} =& \frac{1}{Z_{ij}} \localShort{i} \localShort{j} \pairwiseShort{i}{j} \cdot \nonumber\\
   &  \!\!\! \prod_{\RV{k} \in \neighbors{i} \backslash {j}} \!\!\!\!\!\!\!\! \msg[\circ]{k}{i}{} \cdot  \!\!\! \!\!\! \prod_{\RV{l} \in \neighbors{j} \backslash {i}}  \!\!\!\!\!\!\!\! \msg[\circ]{l}{j}{},
 \label{eq:marginalsPairwise}
\end{align}
where $Z_i,Z_{ij} \in \RealPositive$ guarantee that all probabilities sum to one. 
We refer to the set of singleton and pairwise marginals as pseudomarginals $\pseudomarginals$, where 
\begin{align}
 \pseudomarginals = \{\marginalsShortApprox{x_i},\marginalsShortApprox{x_i,x_j} : \RV{i} \in \setOfNodes, \edge{i}{j} \in \setOfEdges\}.
 \label{eq:pseudomarginals}
\end{align}
We further denote the pseudomarginals obtained at a fixed point of BP by $\pseudomarginalsMinLocal$.\footnote{ 
With slight abuse of notation, we will refer to both the pseudomarginals $\pseudomarginalsMinLocal$ and the messages $\setOfMsg[\circ]$ as fixed points throughout this work.
Strictly speaking, however, the pseudomarginals are only evaluated at a fixed point rather than being a fixed point themselves.}

\subsection{THE BETHE APPROXIMATION \& RELATED WORK}\label{subsec:background:energy}
BP is closely connected to variational methods and concepts from statistical physics. 
In particular, a direct relationship between performing BP and minimizing the Bethe free energy exists (cf.~\cite{yedidia}).

The Bethe free energy $\FB(\pseudomarginals)$
is evaluated over the pseudomarginals (i.e., the singleton- and pairwise marginals) and is a function of the average energy $\EB(\pseudomarginals)$ and the Bethe entropy $\SB(\pseudomarginals)$ that are defined by
\begin{align}
 \EB(&\pseudomarginals)=- \sum_{\RV{i}}\sum_{x_i}\marginalsShortApprox{x_i}\ln\localShort{i} \nonumber\\
 & -\sum_{\edge{i}{j} \in \setOfEdges} \sum_{x_i,x_j}\ \marginalsShortApprox{x_i,x_j} \ln {\pairwiseShort{i}{j}}.\label{eq:energy}\\
 \SB(&\pseudomarginals) = - \sum_{\edge{i}{j} \in \setOfEdges} \sum_{x_i,x_j}\marginalsShortApprox{x_i,x_j} \ln \marginalsShortApprox{x_i,x_j} \nonumber \\
 &+ \sum_{\RV{i}}(|\neighbors{i}|-1)  \sum_{x_i}\! \marginalsShortApprox{x_i}\ln\marginalsShortApprox{x_i}.\label{eq:entropy}
\end{align}
This subsequently defines the Bethe free energy according to 
\begin{align}
 \FB(\pseudomarginals) =& \EB(\pseudomarginals) - \SB(\pseudomarginals). \label{eq:FB}
\end{align}
Let us define the local polytope $\LP$, i.e, the set of consistent pseudomarginals, where 
\begin{align}
 \LP = \{\pseudomarginals : \sum_{x_i} \marginalsShortApprox{x_i} = 1, \sum_{x_j} \marginalsShortApprox{x_i,x_j} = \marginalsShortApprox{x_i}\}. \nonumber
\end{align}
Minimizing $\FB$ over the local polytope gives us the desired global minimum $\FBMinGlobal=\min_{\LP}\FB(\pseudomarginals)$.
This, however, is not straightforward as the constrained $\FB$ is often non-convex.

The main reason for considering the Bethe free energy is its immediate connection with the fixed points of BP $\setOfMsg[\circ]$ (and the associated pseudomarginals $\pseudomarginalsMinLocal$):
all stationary points $\FBMinLocal$ are in a one-to-one correspondence with the fixed points of BP.
More precisely, we have
\begin{align}
 \FBMinLocal = \FB(\pseudomarginalsMinLocal).
\end{align}
In practice, we are particularly interested in stable fixed points (for which BP converges if initialized close enough);
we index all stable fixed points $\pseudoTemplate{}{s}$  by $s = 1,\ldots,S$ and denote the set of all stable fixed points by $\vm{S}$.
Likewise, we consider the set of all fixed points that constitute minima of the Bethe free energy $\vm{M}$ and index them by $m=1,\ldots,M$.
Note that every stable fixed point $\pseudoTemplate{}{s}$ must be a minimum of $\FB$;
the converse, however, need not be the case, i.e., not every local minimum corresponds to a stable fixed point, so that $\vm{S} \subseteq \vm{M}$~\cite{watanabe}.

This correspondence between BP and $\FB$ led to a better understanding of BP and
inspired plenty of methods that minimize $\FB$ directly~\cite{welling2003approximate,cccp2003yuille}.
The minimization, however, is still highly non-trivial and requires good approximation methods in practice. 
Since strong pairwise potentials often reduce the accuracy of the Bethe approximation and are responsible for its non-convexity~\cite{knoll2017fixed}, one can correct the entropy term~\eqref{eq:entropy} by accounting for the strong potentials;
this admits convex relaxations that
provide provable convergent message passing algorithms~\cite{wainwright2003tree, meltzer2009convergent,globerson2007convergent, hazan2008convergent,meshi2009convexifying}. 
There is, however, a trade-off between convergence-properties and accuracy in general.
That is, if it can be minimized, the Bethe approximation often provides accurate results and outperforms convex free energy approximations~\cite{meshi2009convexifying, weller2013approximating}. 
Thus, it is a relevant problem to directly approximate $\FB$ in a way that allows for efficient minimization.
Polynomial run-time algorithms exist that approximate $\FB$ for restricted models: these include sparsity constraints~\cite{shin2012complexity} or require attractive models~\cite{weller2013approximating}. If both properties are fulfilled, i.e., for locally tree-like attractive models the Bethe approximation is exact and can be optimized efficiently~\cite{dembo2010ising}. Note that $\FB$ provides an upper bound on $\FGibbs$ for attractive models~\cite{ruozzi2013bethebound,willsky2008loop,weller2014clamping}.

We aim to estimate $\FB$ and $\pseudomarginalsExact$;
while the approximation of $\FB$ in~\cite{weller2013approximating} is $\varepsilon$-accurate, no run-time guarantees exist for general models.
Our work, on the contrary, provides an estimate in constant run-time (see Theorem~\ref{thm:complexity} in Section~\ref{sec:theory}); the approximation error, however, cannot be made arbitrarily small in general.
Both methods overcome their respective disadvantages when restricting the models; i.e., both methods do efficiently minimize the Bethe approximation for unidirectional models.\footnote{ 
The restriction to unidirectional models is actually too conservative for~\cite{weller2013approximating}, where the approximation of $\FB$ takes polynomial run-time for balanced models as well.}

\subsection{Solution Space of Binary Pairwise Models}\label{subsec:solution_space}
Over the years, a substantial amount of work has been devoted to getting a good understanding of how the solution space is affected by the specification of the graphical model.
We are particularly interested in how the performance of BP and the energy landscape of $\FB$ vary, depending on the specification of the potentials.
Here, we briefly summarize the most important results with a particular focus on how the solution space changes when considering increasingly stronger couplings.

This discussion serves two purposes: 
on the one hand, it provides a good intuition and motivates the working principle of SBP discussed in Section~\ref{sec:SBP};
on the other hand, it prepares the required background for our theoretical analysis in Section~\ref{sec:theory}.
Moreover, the difference in the behavior of BP is often only considered between general and attractive models. 
Instead, we consider the difference in BP's behavior between frustrated, balanced, and unidirectional models. 
We are convinced that this distinction is not only relevant for analyzing SBP in this work but that this is a crucial distinction one must make in the pursuit of understanding what makes some problems inherently hard to solve.\\

For as long as the couplings $\coupling{i}{j}$ are sufficiently small (relative to $\field{i}$), the Bethe free energy $\FB$ is convex, BP has a unique fixed point, and BP converges~\cite{knoll2017fixed,mooij2007sufficient}.
As the couplings increase, the performance tends to suffer as multiple fixed points emerge and as BP may fail to converge (even if the fixed point is still unique)~\cite{mooij2007sufficient, knoll2017uai}.\footnote{ 
Note that the differences in behavior coincide with different phases in statistical mechanics (cf.~\cite{zdeborova2016statistical,mezard2009, tatikonda2002loopy}).}
The precise effect of the strong couplings on the behavior of BP and the shape of $\FB$ mainly depends on the respective model-class.
Thus, we will split the following discussion accordingly.

For \emph{frustrated models}, the number of stationary points scales exponentially with the number of variables.
Often, multiple stationary points coexist with the same value of $\FB$ but possibly completely different marginals;
it may even happen that there is not a single unambiguous global minimum.
Moreover, for frustrated models, the Bethe free energy is not lower bounded by the exact free energy, so that the global minimum need not approximate the free energy best.
Such  complex energy landscapes render any attempt of minimizing the Bethe free energy futile.
Moreover, in particular for models with strong couplings, BP often fails to converge for frustrated models.

For \emph{balanced models}, the number of stationary points still scales exponentially with the number of variables.
For example, attractive models with arbitrary local potentials (i.e., random field Ising models) are some of the simplest forms of disordered systems~\cite{belanger1991random}. 
Yet, in comparison with frustrated models, they are much easier to solve.
Since every balanced model can be transformed into an equivalent attractive model, the Bethe free energy is lower bounded by the exact free energy.
Moreover, many methods that minimize $\FB$ are particularly efficient for attractive models and thus for balanced models as well.
One of the properties that we will use throughout is that for attractive models all fixed points corresponding to local minima of $\FB$ are stable~\cite[Theorem 6]{watanabe},~\cite{perugini,stoop}.
Although BP may still fail to converge on balanced models, it is thus at least guaranteed that BP will converge if the messages are initialized close enough to a fixed point.

For \emph{unidirectional models} two fixed points exist at most.
This not only renders approximating the free energy and the marginals relatively straightforward but also allows for a thorough evaluation of various algorithms.
Since we devote a great part of our theoretical analysis of SBP to its performance on unidirectional models, we will now discuss the solution space of unidirectional models in more detail and prepare some foundations for the later analysis.

Most properties result from generalizing the properties of homogeneous models.
As discussed in Section~\ref{subsec:model:classes} these models qualitatively behave the same.

First, we treat the special case where all local potentials are specified by $\field{i}=0$ and where both fixed points -- if they exist -- are symmetric.
\begin{lm}[]\label{lm:symmetry}
  Let us consider a unidirectional model with loops, with $\field{i} = 0$, and with couplings strong enough to admit multiple (i.e., two) fixed points.
  Then, both fixed points $\stableOne$ and $\stableTwo$ are symmetric, i.e., $\mean{i}^{\stableOne}=-\mean{i}^{\stableTwo}$, and  both fixed points have the same Bethe free energy, i.e.,  $\FB^{\stableOne}= \FB^{\stableTwo}$.
\end{lm}
\begin{IEEEproof}
  Given that the local potentials are the same for both states, the message update rule from~\eqref{eq:update} simplifies to $\msg[n+1]{i}{j}{} \propto \sum_{x_i}   \pairwiseShort{i}{j}  \prod_{X_k} \msg[n]{k}{i}{}$.
  Consequently, if $\msg[\stableOne]{i}{j}{} = \msg[\circ]{i}{j}{}$ are fixed point messages, so are the symmetric messages $\msg[\stableTwo]{i}{j}{} = 1-\msg[\circ]{i}{j}{}$.
  By computing the pseudomarginals according to~\eqref{eq:marginalsSingle}-\eqref{eq:marginalsPairwise} it follows immediately that the means are symmetric around 0 and that both fixed points have the same Bethe free energy (see Example~\ref{ex:vanishing}).
\end{IEEEproof}

We will now present an illustrative example that describes the behavior of models with $\field{}=0$.
These models have been well-studied in the literature~\cite{mooij2005properties, weller2014understanding, knoll2017fixed}.
We present Example~\ref{ex:vanishing} to summarize the most important properties that serve as the basis for our theoretical analysis of SBP in Section~\ref{sec:theory}.
\begin{ex}\label{ex:vanishing}
Consider a homogeneous model on a complete graph with $N=4$ variables, positive couplings $\coupling{i}{j} = \coupling{}{} > 0$, and local potentials specified by $\field{i}=0$.
For computing the Bethe free energy $\FB$ according to~\eqref{eq:FB}, we enforce the symmetry of the model by setting all singleton marginals to the same value $\marginalsShortApprox{x_i} =\marginalsApprox{}$.
We compute the Bethe-optimal pairwise marginals according to~\cite[Lemma 1]{welling2001belief} (see also~\cite{weller2014understanding}). 
This allows us to compute the pairwise marginals given only the singleton marginals.\footnote{ 
Note that an alternative representation of~\eqref{eq:joint} is used in~\cite{welling2001belief, weller2014understanding}; there is, however, a straightforward mapping from one representation to the other.}

The effect of the coupling strength is illustrated in Fig.~\ref{fig:energy:nolocal}.
Note that for $\field{}=0$,  $\marginalsApprox{}=(0.5,0.5)$ is a fixed point for every value of $\coupling{}{}$; we will often refer to this fixed point in terms of its means $\mean{i}=0$.\footnote{ 
This fixed point exists whenever all local potentials are specified by $\field{}=0$ and can be computed analytically~\cite{mooij2005properties}.}
Consequently, there is also a corresponding stationary point of $\FB$.
For small values of $\coupling{}{}$, $\FB$ is convex and this fixed point is unique (see Fig.~\ref{fig:energy:local1});
as $\coupling{}{}$ increases, $\FB$ becomes flat (see Fig.~\ref{fig:energy:local2}), before turning into a local maximum with two symmetric minima branching off (see Fig.~\ref{fig:energy:local3}).

\begin{figure}[t]
\centering
\centering
	\subfloat[][]{\label{fig:energy:nolocal1} \includegraphics[width=0.3\linewidth]{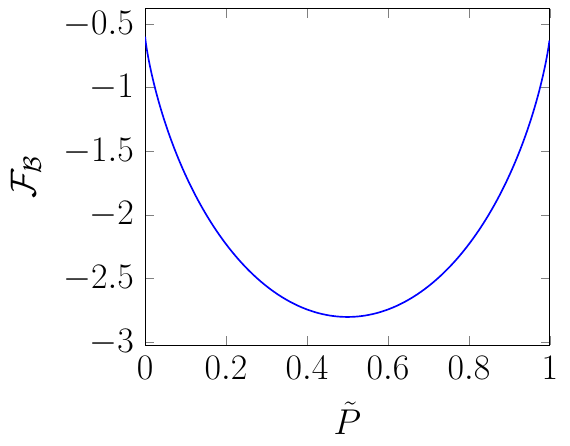}}
	\subfloat[][]{\label{fig:energy:nolocal2} \includegraphics[width=0.3\linewidth]{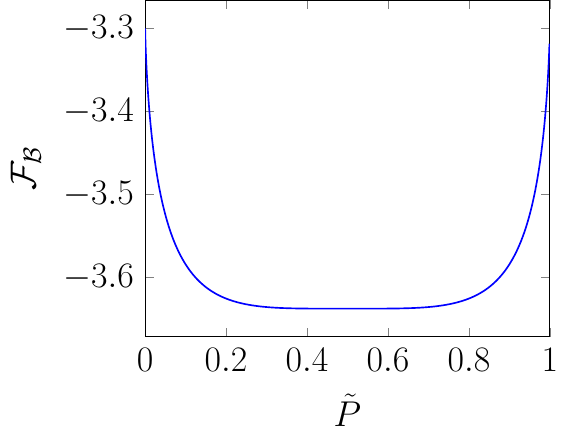}}
	\subfloat[][]{\label{fig:energy:nolocal3} \includegraphics[width=0.3\linewidth]{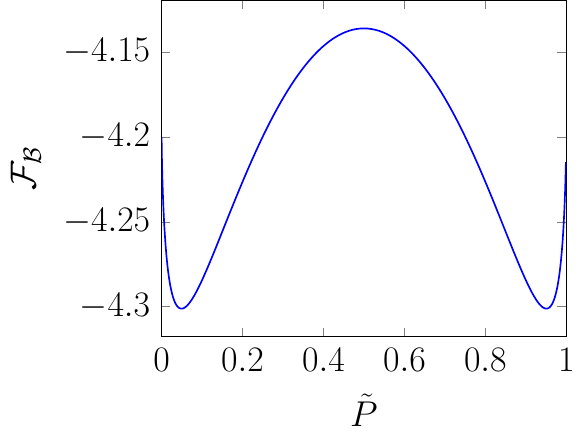}}
	\caption{Evolution of the Bethe free energy $\FB$ for Example~\ref{ex:vanishing}.
	We consider a complete graph with $N=4$ variables with homogeneous potentials, attractive edges, and $\field{i}=0$.
	$\FB$ is evaluated for $\marginalsShortApprox{x_i} =\marginalsApprox{}$ for
	(a) $\coupling{}{}=0.1$, (b) $\coupling{}{}=0.55$, and (c) $\coupling{}{}=0.7$.}
	\label{fig:energy:nolocal}
\end{figure}
\end{ex}

Second, when all local potentials are specified by $\field{i}>0$ the fixed points are not symmetric anymore. 
For later reference, we summarize some important properties of the fixed points in the following Lemma.
\begin{lm}[]\label{lm:symmetry:not}
 Let us consider a unidirectional model with $\field{i}>0$ and couplings strong enough to admit multiple (i.e., two) fixed points.
 Then, one fixed point has its means aligned with the local potentials, i.e., $\mean{i}^\stableOne > 0$;
 we call this fixed point consistent or state-preserving (cf.~\cite{goldberg2003computational, knoll_accuracy}).
 Similar, as in Lemma~\ref{lm:symmetry}, the means of the second fixed point have opposing signs, i.e., $\mean{i}^\stableTwo < 0$.
 Both fixed points are not symmetric, however, and the consistent fixed point has larger means $|\mean{i}^{\stableOne}| > |\mean{i}^{\stableTwo}|$ and larger correlations $\correlation{i}{j}^{\stableOne} > \correlation{i}{j}^{\stableTwo}$.
 Accordingly, the consistent fixed point constitutes the global minimum of the Bethe free energy, i.e., $\FBMinGlobal = \FB^{\stableOne}$.
\end{lm}
\begin{IEEEproof}
First, the existence of a consistent fixed point, i.e., $\mean{i}^\stableOne > 0$, is an immediate consequence of bounding the location of the fixed points of BP in~\cite[Theorem 3]{weller2013bethe},\cite{yedidia}.
Second, the existence of a second point with opposing means, i.e., $\mean{i}^\stableTwo < 0$, follows from the fact that only a single fixed point exists with positive means~\cite[Lemma 2.3]{dembo2010ising}; 
thus the second fixed point, if it exists, must have negative means (see Example~\ref{ex:unidirectional}).
Third, it is an immediate consequence of the update rule in~\eqref{eq:update} that the consistent fixed point is energetically favored, i.e., $\msg[s]{i}{j}{} > \msgShort[t]{i}{j}{-x_j}$.
Then, computing the marginals according to~\eqref{eq:marginalsSingle} implies that the consistent fixed point has larger means, i.e., $|\mean{i}^{\stableOne}| > |\mean{i}^{\stableTwo}|$.
The same line of reasoning shows that the correlations are larger as well.
Finally, the consistent fixed point has a lower average energy, i.e., $ \EB^s(\pseudomarginals) <  \EB^t(\pseudomarginals)$ (see~\eqref{eq:energy}).
Therefore (and by symmetry of the Bethe entropy), it follows that $\FBMinGlobal = \FB^{\stableOne}$.
\end{IEEEproof}

\begin{ex}\label{ex:unidirectional}
Consider a unidirectional homogeneous model on a complete graph with $N=4$, positive couplings $\coupling{i}{j} = \coupling{}{} > 0$, and local potentials specified by $\field{i} = \field{} = 0.02$.
Again, we enforce symmetry by setting all singleton marginals to the same value and compute the pairwise marginals and the Bethe free energy accordingly.

Similar as in Example~\ref{ex:vanishing}, we illustrate the effect of the coupling strength on the shape of $\FB$.
In some sense we see a similar picture emerging; i.e., $\FB$ is convex for small values of $\coupling{}{}$ (see Fig.~\ref{fig:energy:local1}) and non-convex for larger values of $\coupling{}{}$ (see Fig.~\ref{fig:energy:local3}).\footnote{
Contrary to Example~\ref{ex:vanishing}, no closed-form solution exists for computing the fixed points because of the local potentials.}
There is one crucial difference however:
because of $\field{i}>0$ the minimum does not turn into a maximum with two minima branching of.
Instead, we see that the global minimum persists for all values of $\coupling{}{}$, whereas the energy landscape becomes flat before two additional stationary points emerge somewhere else (see Fig.\ref{fig:energy:local2}).

\begin{figure}[t]
\centering
\centering
	\subfloat[][]{\label{fig:energy:local1} \includegraphics[width=0.3\linewidth]{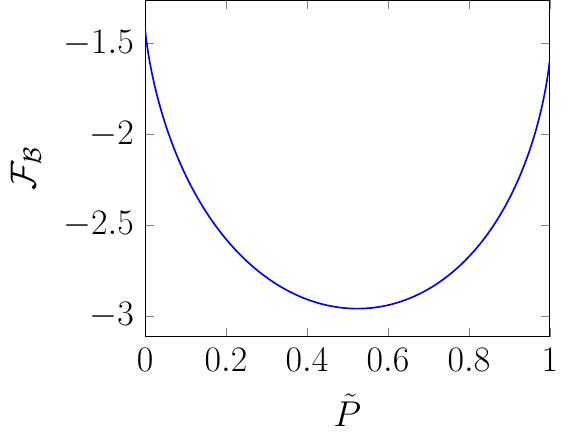}}
	\subfloat[][]{\label{fig:energy:local2} \includegraphics[width=0.3\linewidth]{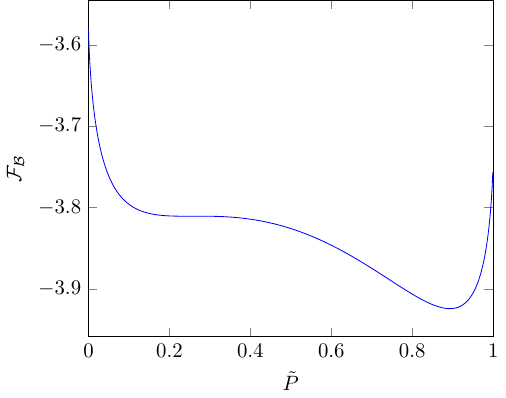}}
	\subfloat[][]{\label{fig:energy:local3} \includegraphics[width=0.3\linewidth]{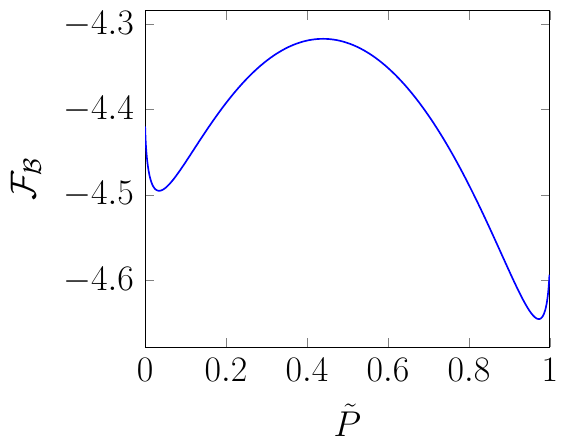}}
	\caption{Evolution of the Bethe free energy $\FB$ for Example~\ref{ex:unidirectional}.
	We consider a complete graph with $N=4$ variables with homogeneous potentials, attractive edges, and $\field{i}>0$.
	$\FB$ is evaluated for $\marginalsShortApprox{x_i} =\marginalsApprox{}$ for
	(a) $\coupling{}{}=0.25$, (b) $\coupling{}{}=0.62$, and (c) $\coupling{}{}=0.75$.}
	\label{fig:energy:local}
\end{figure}
\end{ex}
We will see in Section~\ref{sec:theory} that unidirectional (non-homogeneous) models behave qualitatively the same;
consequently we will make repeated use of the behaviour illustrated in Example~\ref{ex:unidirectional} for analysing the behaviour of SBP.

Finally, note that every fixed point that corresponds to a local minimum is a stable fixed point of BP for unidirectional models. This is a direct consequence of~\cite[Theorem 6]{watanabe} and the fact that unidirectional models are attractive per definition.
Even more importantly, BP will always converge for unidirectional models: 
therefore note that BP always converges for unidirectional models with $\field{i}=0$~\cite{mooij2007sufficient} and that the existence of local potentials with  $\field{i} \neq 0$ can only enhance the convergence properties~\cite{knoll2017uai}. 
Recently, it has also been shown that for unidirectional models BP will converge reasonably fast, if initialized properly~\cite{koehler2019fast, dembo2010ising}.

\subsubsection{Combination of Fixed Points}
If multiple fixed points exist, the performance of BP often varies considerably between different fixed points.
Here we briefly introduce the RSB (replica symmetry breaking) assumption that states how all those fixed points can be combined to compute the exact marginals.
Therefore, let us consider all local minima of the Bethe free energy $\FB^m$, the pseudomarginals $\pseudoTemplate{}{m}$ at the corresponding fixed point, and the associated partition function approximation $\PartitionApprox^m$.

\begin{con}[RSB Assumption]\label{lm:rsb}
 Let us consider all $M$ local minima of the Bethe free energy $\FB^m$, the pseudomarginals at the corresponding fixed point $\pseudoTemplate{}{m}$, and the associated partition function approximation $\PartitionApprox^m$.
 Then, the exact singleton marginals $\marginalsShort{x_i}$ are given by a convex combination of all fixed points at minima of $\FB$ according to
 \begin{align}
  \marginalsShort{x_i} = \frac{1}{\sum\limits_{m=1}^M \PartitionApprox^m} \sum_{m=1}^{M}\tilde{P}^m(x_i) \PartitionApprox^m.
  \label{eq:rsb}
 \end{align}
\end{con}
This representation belongs to a set of postulates in~\cite{mezard1987spin}.
One underlying assumption is that the system does exhibit multiple fixed points (unique fixed points would falsely imply exact marginals otherwise).
Despite its non-rigorous flavor, Conjecture~\ref{lm:rsb} has been verified for a wide range of problems~\cite[Ch.19]{mezard2009}.
Example~\ref{ex:vanishing} and Fig.~\ref{fig:energy:nolocal3} in particular give an intuitive illustration of the RSB assumption: for this model the local maximum (i.e., the unstable fixed point) corresponds to the exact marginals; yet, both local minima do not approximate the marginals well. 
Combining both minima according to~\eqref{eq:rsb}, however, gives the exact singleton marginals because of the model's symmetry properties.

Obtaining all minima of the Bethe free energy, however,  is a complex task that is only possible for models with certain structures or potentials~\cite{coja2019bethe, zdeborova2016statistical,knoll2017fixed, perugini}.
Besides the application to optimization problems~\cite{braunstein2005survey} the RSB assumption is thus still of limited practical relevance for estimating the marginals; nonetheless, it provides a powerful concept for assessing and comparing the marginals given a selection of fixed points~\cite{knoll_accuracy}.

\section{Self-guided belief propagation (SBP)}\label{sec:SBP}
The main aim of SBP is to select an accurate fixed point, i.e., a fixed point that approximates the singleton marginals well, or -- if none are stable -- to approximate an accurate fixed point.
In this section we present an intuitive justification of the proposed method and subsequently introduce SBP in detail. 
We further present practical considerations and pseudocode of SBP.  
A formal treatment of SBP is presented in Section~\ref{sec:theory}.

The current understanding of BP is that strong pairwise potentials negatively influence BP in the sense of deteriorating the convergence behavior~\cite{mooij2005properties, knoll2017fixed} and the approximation quality of the marginals~\cite{knoll2017fixed,knoll_accuracy,weller2014understanding}
and that incorporating the potentials slowly~\cite{braunstein2007encoding} may reduce the overall number of iterations.
Inspired by the recent observation that strong local potentials increase accuracy and lead to better convergence properties~\cite{knoll2017uai}, we aim to reduce the influence of the pairwise potentials that negatively influence BP. 
In doing so, we hope that an accurate fixed point emerges if we start from a simple model with independent random variables and slowly increase the potentials' strength.

SBP starts from a simple model with independent random variables and slowly incorporates the edge potentials' strength.
More precisely, SBP is a homotopy method that solves the simple problem first and -- by repetitive application of BP -- keeps track of the fixed point as the interaction strength is increased by a scaling term.

Formally, SBP considers an increasing length-$M$ sequence $\{\scaling_m\}$ where $m=1,\ldots,M$ such that $\scaling_{m} < \scaling_{m+1}$ and $\scaling_m \in [0,1]$ with $\scaling_1 = 0$ and $\scaling_M = 1$.
This further indexes a sequence of probabilistic graphical models $\{\PGM[m]\}$ that converges to the model of interest $\PGM[M] = \PGM$. 
Every probabilistic graphical model has a set of potentials $\setOfPot_{m} = \{\pairwiseSBP{i}{j}{m},\localSBP{i}{m}\}$ associated, where $\localSBP{i}{m} = \localShort{i}$ and the pairwise potentials at index $m$ are exponentially scaled by
\begin{align}
  \pairwiseSBP{i}{j}{m} &=   \exp(\coupling{i}{j}\scaling_m x_i x_j)\nonumber \\
  &= \pairwiseShort{i}{j}^{\scaling_m}.
  \label{eq:scaling:potentials}
\end{align}
We further write $\setOfMsg[\circ]_\iteration{m}$ to clarify that we consider a BP fixed point for the model $\PGM[m]$.
If multiple fixed points exist, the initialization of BP plays a crucial role in determining to which fixed point BP converges and how it performs.
SBP always provides a favorable initialization by the preceding fixed point and aims to perform the composite function\footnote{
The underlying assumption is the existence of a continuous path that connects the initial messages $\setOfMsg[\init]_{\iteration{1}}$ to the final fixed point $\setOfMsg[\circ]_\iteration{M}$, where all intermediate initializations $\setOfMsg[\init]_\iteration{m}$ lie along this path. For now, we assume a fixed number of models $M$ for practical reasons but we will consider SBP's behavior in the limit of $M\rightarrow\infty$ in our formal analysis in Section~\ref{sec:theory}.}
\begin{align}
  \setOfMsg[\circ]_\iteration{M} = \BP^\circ_\iteration{M}\left( \BP^\circ_\iteration{M-1}\left( \cdots \BP^\circ_\iteration{1}\left( \setOfMsg[\init]_{\iteration{1}} \right)   \right)\right).
  \label{eq:BP_composition}
\end{align}
This may lead to problems if the fixed point becomes unstable for some value $m<M$, in which case we cannot rely on BP to keep track of the fixed point anymore. 
Instead, SBP provides the last stable fixed point in that case, i.e., $\setOfMsg[\circ]_\iteration{m}$ as the final estimate.

\begin{figure}[t]
 \centering
 \includegraphics[width = 0.95\linewidth]{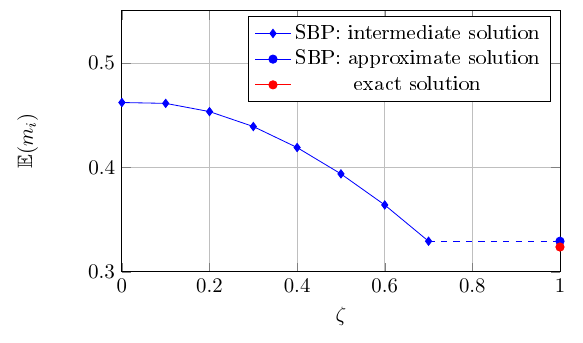} 
 \caption{Example~\ref{ex:solPath}: SBP proceeds along the smooth solution path and obtains accurate marginals despite instability of the terminal fixed point.}
 \label{fig:example}
\end{figure}

\begin{ex}\label{ex:solPath}
	We illustrate how SBP approximates the marginals for a frustrated model, for which BP fails to converge, in Fig.~\ref{fig:example}.
	Initially, SBP obtains the pseudomarginals for $\scaling=0$ by running BP on the simple model. 
	As $\scaling$ increases, SBP estimates the pseudomarginals along the solution path by running BP to keep track.
	For $\scaling > 0.7$, however, BP does not converge anymore; 
	SBP consequently terminates and provides the marginals of the last stable fixed point as the final approximate solution.
	Note that the approximated marginals are already close to the exact ones in this example; experiments show that this is often the case (see Section~\ref{sec:experiments}).
\end{ex}

In other words, SBP relaxes the problem of minimizing $\FB$ by making all variables independent (and the Bethe approximation exact). 
Then the problem is deformed into the original one by increasing $\scaling$ from  zero to one. 
Thereby, a stationary point $\FBMinLocal$ emerges as a well-behaved path (see Property~\ref{prop:properties} in Section~\ref{sec:theory}) and SBP keeps track of it with BP constantly correcting the stationary point.

\subsection{Practical considerations}\label{subsec:sbp:refinements}
In practice the run-time of SBP is influenced by the difference between two successive fixed points $\setOfMsg[\circ]_{m}$ and $\setOfMsg[\circ]_{m-1}$ -- 
the difference is primarily determined by the number of steps $M$.
Ideally $M$ should be as large as possible. This, however, increases the run-time (see Theorem~\ref{thm:complexity}); in practice, we would choose $M$ as small as possible but as large as necessary.
Moreover, one can adaptively increase the step size if two successive fixed points are close, i.e., if $\setOfMsg[\circ]_{m}\simeq \setOfMsg[\circ]_{m-1}$ (cf.~\cite[pp.23]{sommese2005numerical},~\cite{allgower2003numerical}). 
Our experiments show that 
it is sufficient to use rather coarse steps; we used $M\leq 10$ for all reported experiments and using more steps would not have improved the accuracy.

Additionally, instead of initializing $\BP_{m}$ with its preceding fixed point messages, i.e.,

$\setOfMsg[\text{init}]_{m} = \setOfMsg[\circ]_{m-1}$ one can (e.g., by spline extrapolation)  estimate $\setOfMsg[\init]_{m} = f(\setOfMsg[\circ]_{m-1},\setOfMsg[\circ]_{m-2},\ldots,\setOfMsg[\circ]_{m-k})$ so that $\setOfMsg[\init]_{m} \cong \setOfMsg[\circ]_{m}$ to reduce the overall number of iterations.
We empirically observed that the benefit diminishes for $k > 3$.

\subsection{Pseudocode} \label{subsec:sbp:pseudocode}
  \input{SBP_pseudocode}
\section{Experiments}\label{sec:experiments}
We apply SBP to  attractive (Section~\ref{subsec:experiments:attractive}) and general (Section~\ref{subsec:experiments:general}) models on $n \times n$ grid graphs of different sizes, on complete graphs with $N=10$ random variables, and on random graphs (i.e., Erdos Renyi graphs with an average degree $\hat{d}=3$) with  $N=10$ random variables.
These graphs are considered in order to render the computation of the exact marginals feasible and to make the results comparable to previous work~\cite{weller2014understanding,sontag2008new,meshi2009convexifying,srinivasa2016survey}.
\begin{figure*}[!t]
	\subfloat[][]{\label{fig:attractive} \includegraphics[width=0.5\linewidth]{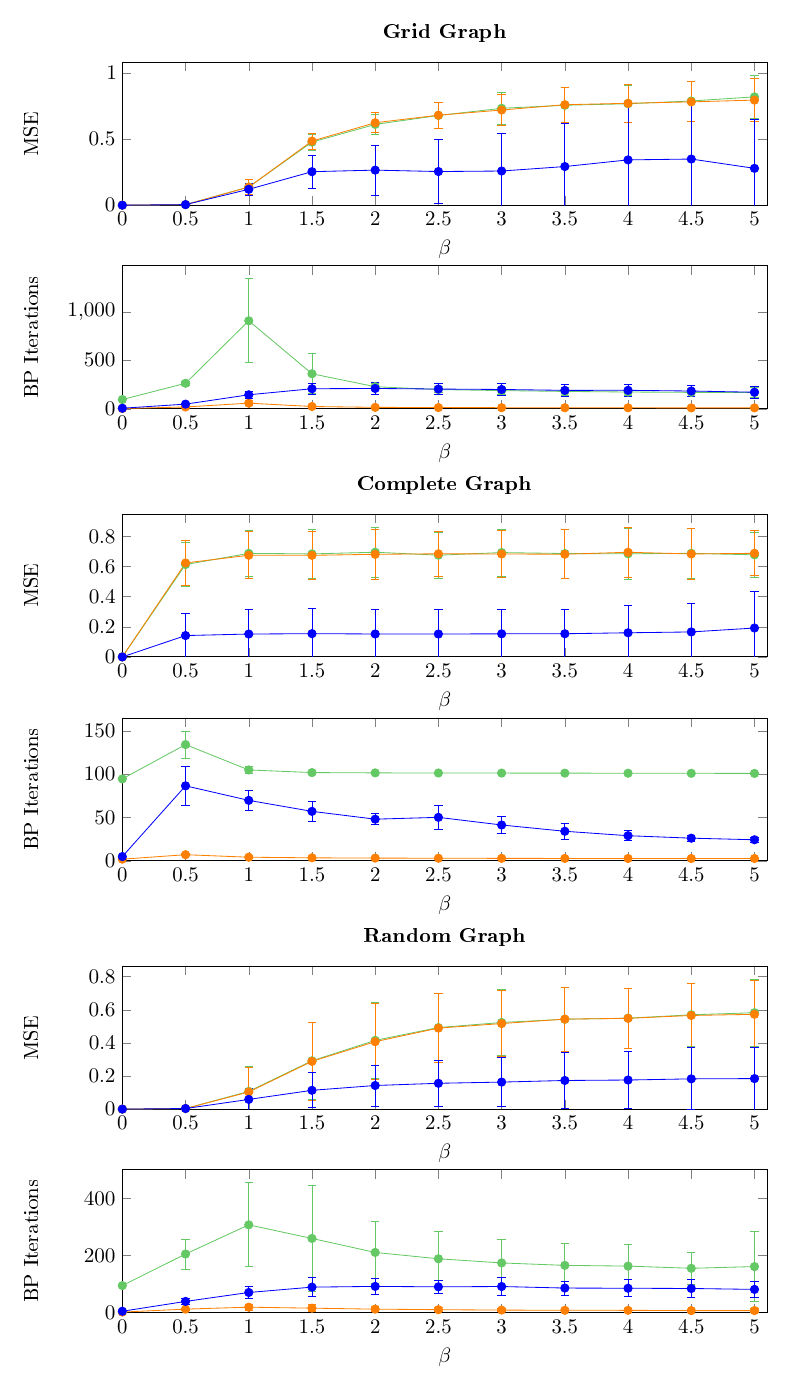}}
	\subfloat[][]{\label{fig:general} \includegraphics[width=0.5\linewidth]{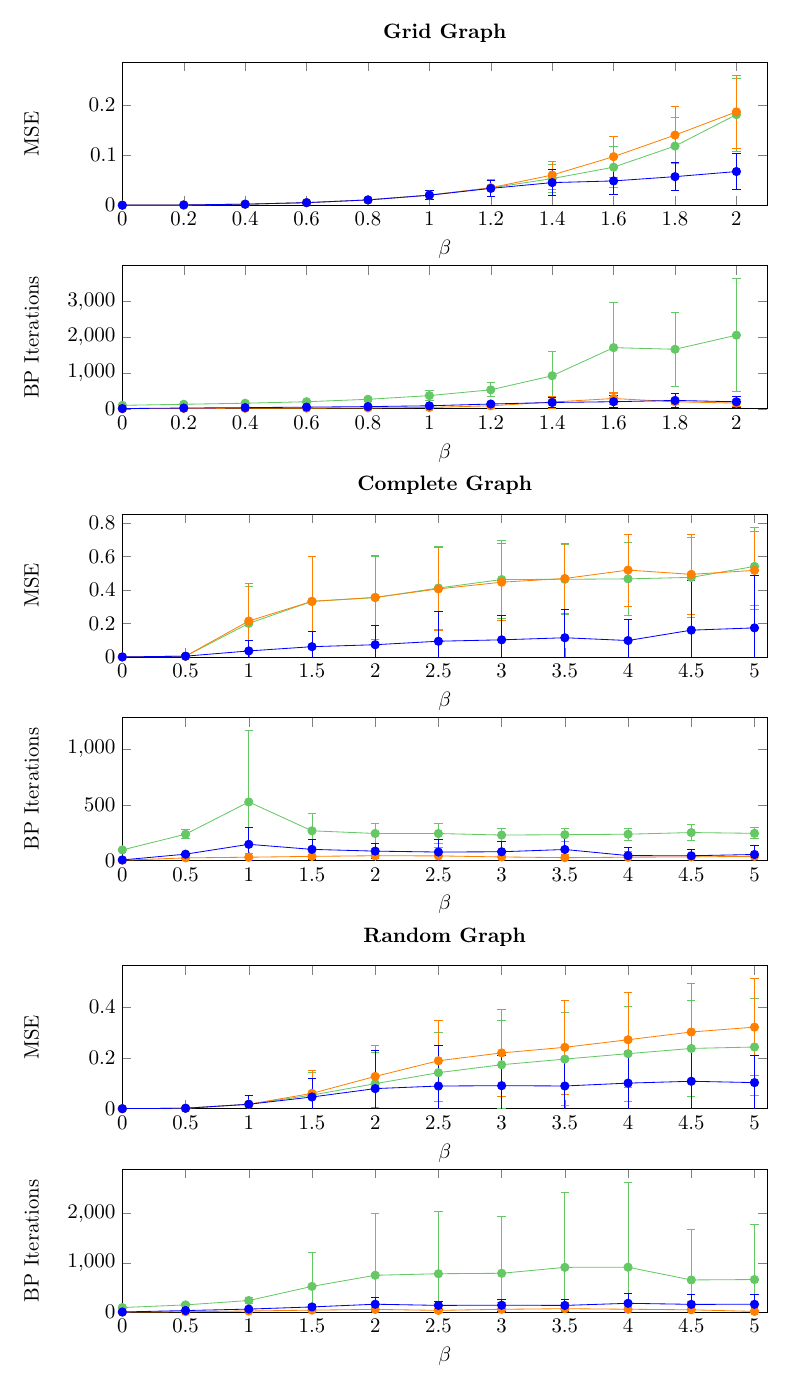}}	
	\caption{$\mse$ and number of iterations for: $ \SBP_{all} $ (blue), $ \text{BP}^\circ $ (orange), and $ \BPD^\circ $ (green);	
		$\field{i} \sim \mathcal{U}(-0.5, 0.5)$ and (a) $\coupling{i}{j} \sim \mathcal{U}(0,\beta)$ (attractive model); (b) $\coupling{i}{j} \sim \mathcal{U}(-\beta,\beta)$ (general model). 
		In terms of accuracy, SBP is superior in all scenarios, while increasing the number of iterations only slightly.}
\end{figure*}
\subsection{Experimental settings}\label{subsec:experiments:evaluation}

\begin{table*}[!t]
  \caption{\textsc{Results for general models with $\coupling{i}{j} \in \{-1,1\}$ on grid graphs ($ N \!=\! 25$ and $ N \!=\! 100$), complete graphs ($ N \!=\! 10$), and random graphs ($ N\!=\!10$). 
  We report the MSE to the exact marginals and the $\mseb$ to the Bethe approximation, convergence ratio, and the overall number of BP iterations.
  Only converged runs are considered for $ \text{BP}^\circ $ and $ \BPD^\circ$ but all runs are considered for $\text{SBP}_{all}$, $\text{Gibbs}_{all}$, and $\FB[all]^*$.}}
  \label{table:general}
    \centering
  \footnotesize
  \begin{center}
	  \begin{tabular}{llcccccccccccc} 
	  \toprule
    \multicolumn{2}{c}{}& \multicolumn{3}{c}{Grid Graph( $5\times5$)} & \multicolumn{3}{c}{Grid Graph ($10\times10$)} & \multicolumn{3}{c}{Complete Graph} & \multicolumn{3}{c}{Random Graph} \\
    \cmidrule(lr){3-5} \cmidrule(lr){6-8} \cmidrule(lr){9-11} \cmidrule(lr){12-14}
	  &$\mathbf{\theta}$	&  ${0}$ & ${0.1} $  & ${0.4} $ & ${0} $ & ${0.1} $  & ${0.4} $& ${0} $ & ${0.1} $  & ${0.4} $ & ${0} $ & ${0.1} $  & ${0.4} $  \\ 
  \midrule
	  
  \multirow{5}{*}{$\mse$ }		& $\text{BP}^\circ$  & 0.338 & 0.251 & 0.102		& -     & - & 0.184 &   0.463  & 0.466 & 0.356 & 0.252     & 0.202 & 0.101 \\
  & $\BPD^\circ$  & 0.226 & 0.198 & 0.066 	& 0.186 & 0.240 & 0.154 &  0.463 & 0.473 &  0.422 & 0.128 & 0.116 & 0.083  \\
	  & $\SBP_{all}$  & \textbf{0.000}& 0.029 & \textbf{0.047}& \textbf{0.000}& \textbf{0.026} & \textbf{0.077}& \textbf{0.000}& \textbf{0.055}  & \textbf{0.074}  & \textbf{0.000}& 0.048 & 0.049  \\
	  & $\FB[all]^*$ &  0.036    &  0.042    &  0.069& - & - & - & -& - & -& - & - &- \\
	  & $\text{Gibbs}_{all}$ & 0.001 & \textbf{0.016}    &   0.064& 0.001 & 0.037 &   0.120 &  0.096 &  0.096  & 0.077  & 0.001 & \textbf{0.011} & \textbf{0.048}   \\  \midrule
	  
  \multirow{2}{2cm}{Convergence ratio} & $\text{BP}^\circ$  & 0.05    & 0.11    & 0.26  & 0.00    & 0.00    &  0.02   &   0.41    &   0.42   &   0.50   & 0.30     & 0.33    &   0.49 \\
	  & $\BPD^\circ$ & 0.11     & 0.16    & 0.69 & 0.01     & 0.02    &  0.12  &  0.41    &   0.41  &   0.50   & 0.62     & 0.64    &   0.80    \\ \midrule
	  
  \multirow{4}{2cm}{Number of iterations}  & $\text{BP}^\circ$   &    40  &  52  &  84 	&    -  &  -  &  102  &  17     &  17   &    18  	&    42  &  53  &   50   \\
  & $\BPD^\circ$ &   1370   &  1449  &  1735 &   2711  &  2313  &  2599 &   211    &  207   &   234 	&   1077   &  1057  &  873   \\
	  &$\SBP_{all}$&   5   &  182  &   146&   5   &  149  &   209  &   5   &   51  &    110 	&   5   &  149  &  131  \\
	  &$\text{Gibbs}_{all}$& $10^5$ &  $10^5$  &   $10^5$& $10^5$ &  $10^5$  &   $10^5$  &  $10^5$   & $10^5$  & $10^5$	&$10^5$ & $10^5$ & $10^5$\\ \midrule
	  $\mseb$ &    $\SBP_{all}$  & 0.036 & 0.037 & 0.022 & - & - & - & -& - & -& - & - &-\\ 
	  $ \FBMinLocal(\scaling_M) $ equals& $\SBP $& 100 & 10 & 23& - & - & - & -& - & -& - & - &- \\
	  \bottomrule
	  \end{tabular}
  \end{center}
\end{table*}
We evaluate SBP and compare it to BP, $\BPD$ (BP with damping), and Gibbs sampling (run for $10^5$ iterations) using the mean squared error (MSE) between the approximate marginals $\marginalsApprox{\RV{i}}$ and the exact marginals $\marginals{\RV{i}}$, where, for binary random variables,  $\mse = \frac{2}{N} \sum_{i=1}^{N} |\marginals{X_i}(+1) - \marginalsApprox{X_i}(+1)|^2$.
We further compare the run-time by counting the overall number of BP iterations and the number of iterations for Gibbs sampling.\footnote{
Computing the acceptance-probability requires  similar run-time as one BP message update.}

For BP and SBP we set the maximum number of iterations to $N_{BP} = 10^3$ and use random scheduling. 
We randomly initialize $\setOfMsg[\init]$ and use an adaptive step size with an initial step size of $step_{init} = 0.1$ and a threshold of $\varepsilon = 10^{-3}$.
The overall number of steps is thus $ M\leq 10 $.

For $\BPD$ we choose a large damping factor $\epsilon=0.9$ in order to prioritize convergence over run-time and therefore increase the maximum number of iterations to $N_{BP}=10^4$. 
Carefully selecting a damping factor that depends on a given model may reduce the number of iterations until convergence but cannot increase the accuracy; moreover, if chosen too small, $\BPD$ may fail to converge at all.

The initial messages are randomly initialized 100 times for each model, before applying BP with and without damping.
We consider BP (and $\BPD$) as converged for a model if at least a single message initialization (out of 100) exists for which BP converges.
The convergence ratio is the number of experiments (or probabilistic graphical models) for which BP converged at least once divided by the overall number of experiments (i.e., $100$).


The reported error (MSE) and the number of iterations are averaged over all convergent runs of BP and $\BPD$ (i.e., BP\textsuperscript{$\circ$} and $\BPD^\circ$) while all runs that did not converge are discarded.
On the contrary, we average the error and the number of iterations over all models for SBP ($\SBP_{all}$), Gibbs sampling ($\text{Gibbs}_{all}$), and for minimization of the Bethe approximation ($\FB[all]^*$).

\subsection{Attractive models}\label{subsec:experiments:attractive}
We consider grid graphs with $N=10 \times 10$ random variables, random graphs with $N=10$ random variables, and complete graphs with $N=10$ random variables. 
We generate $L=100$ models for every value of $\beta \in \{0,0.5,\ldots,5\}$ and sample the potentials according to $\field{i} \sim \mathcal{U} (-0.5, 0.5)$ and $\coupling{i}{j} \sim \mathcal{U} (0,\beta)$; i.e., overall we consider 1100 different models for each individual graph-structure.
Note that BP is randomly initialized 100 times for every considered model.
We compute the MSE for every value of $\beta$ and visualize the mean and the standard deviation of the $\mse$\footnote{
Note that the MSE is not Gaussian distributed but we report the standard deviation for simplicity.} 
as well as the number of iterations in Fig.~\ref{fig:attractive}.

BP (orange) converges rapidly for all graphs considered; 
hence, there is no additional benefit in using $\BPD$ (green) that only increases the number of iterations.
SBP (blue) slightly increases the number of iterations as compared to BP but converges in fewer iterations than $\BPD$. 
Note that SBP is guaranteed to capture the global optimum when applied to unidirectional models (see Theorem~\ref{thm:accuracy:fb}-\ref{thm:accuracy:marginals}). 
But even if we do allow for random local potentials, we empirically observe that SBP consistently outperforms BP with respect to accuracy; specifically for models with strong couplings
These models exhibit multiple stable fixed points~\cite{knoll2017fixed} such that, depending on the initialization, BP often converges to inaccurate fixed points while SBP selects an accurate one.

\subsection{General models}\label{subsec:experiments:general}
General models with random potentials are often frustrated and traditionally pose problems for BP and other methods that aim to minimize the Bethe approximation.

First, in order to evaluate the performance of SBP we consider uniform local potentials $\field{i} = \field{} \in \{0,0.1,0.4\}$ and draw the couplings with equal probability from $\coupling{i}{j} \in \{-1,1\}$; the results are summarized in Tab.~\ref{table:general}.
Although BP and $\BPD$ fail to converge for most models we observe that SBP stops after only a few iterations and significantly outperforms BP in terms of accuracy. In fact, SBP achieves accuracy competitive with Gibbs sampling but requires three orders of magnitude fewer iterations.

Second, we apply SBP to general graphs and evaluate whether SBP provides a good approximation of the pseudomarginals $\pseudomarginalsMinGlobal$ that correspond to the global minimum of the Bethe free energy $\FBMinGlobal$.
Therefore we consider grid graphs (of size $5\times 5$), which still allows us to approximate the global minimum $\FBMinGlobal$ -- and the related pseudomarginals $\pseudomarginalsMinGlobal$ -- reasonably well by~\cite{weller2013approximating}. The results are summarized in Tab.~\ref{table:general} and show that SBP approximates $\pseudomarginalsMinGlobal$ within the accuracy of our reference method ($\mseb$).
We further report the number of times where SBP obtains the terminal fixed point, i.e., for $\PGM[M]$, in Tab.~\ref{table:general} $i.e., \bigl( \FBMinLocal(\scaling_M)$ equals SBP $\bigr)$.
It becomes obvious that SBP  approximates the terminal fixed point reasonably well, despite frequently stopping for  $\scaling_m < 1$. 
Moreover, the MSE reveals that SBP does not only approximate the pseudomarginals $\pseudomarginalsMinGlobal$ well but concurrently provides an accurate approximation of the exact marginals $\pseudomarginalsExact$.

\begin{figure}[t]
  \centering
  \includegraphics[width=0.95\linewidth]{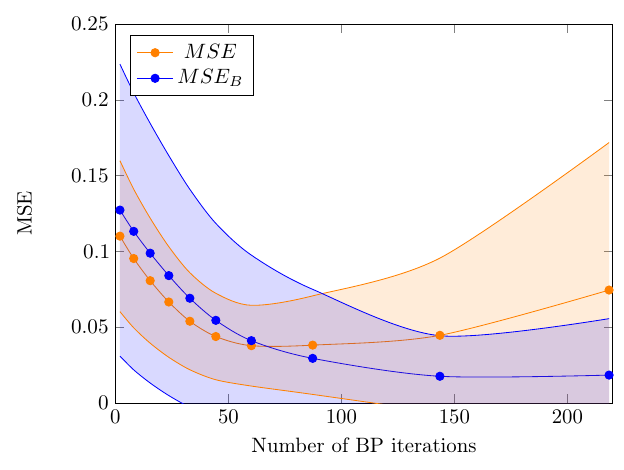}
  \caption{MSE (orange) and $ \mseb $ (blue) and their standard deviation (shaded area) over the cumulative number of iterations. Results are averaged over 100 grid graphs ($5 \times 5$); $\field{i} = 0.4$ and $\coupling{i}{j} \in \{-1,1\}$.}
  \label{fig:mse_iterations}
\end{figure}
Third, we investigate how the approximation quality depends on the scaling term $\scaling_m$. 
Therefore, we depict the evolution of $\mseb$ (to the approximate solution) and of the MSE (to the exact solution) in Fig.~\ref{fig:mse_iterations}.
We observe that $\mseb$ (blue) decreases monotonically with every iteration, which empirically verifies that SBP proceeds along a well-behaved solution path (see Property~\ref{prop:properties}).
 Note that $\mseb$ decreases rapidly in the first iterations and SBP spends a major part of the overall run-time for slight improvements. 
 The MSE to the exact solution, on the other hand, decreases first until it increases again as SBP incorporates stronger couplings. 
 Stronger couplings tend to degrade the quality of the Bethe approximation in loopy graphs and lead to marginals that are increasingly biased towards one state~\cite{weiss2000correctness,knoll2017fixed}.
 This explains why the MSE to the exact solution increases as SBP converges towards the terminal fixed point.
 One could exploit this behavior and restrict the run-time by stopping SBP after consumption of a fixed iteration budget; this may even increase the accuracy with respect to the exact solution.

Finally, we  investigate the influence of the coupling strength: therefore we consider $\field{i} \sim \mathcal{U} (-0.5, 0.5)$ and $\coupling{i}{j} \sim \mathcal{U} (-\beta,\beta)$. 
For every $\beta \in \{0, 0.5, \ldots,5\}$ we consider $L=100$ models and present the averaged results in Fig.~\ref{fig:general}. 
We restrict the results to $\beta \leq 2$ on the grid graph because BP did only converge sporadically for models with stronger couplings.
Even for models where BP converged, SBP requires only slightly more iterations than BP and fewer than $\BPD$.
The benefits of SBP become increasingly evident as the coupling strength increases. 
Again SBP (blue) significantly  outperforms BP\textsuperscript{$\circ$} (orange) and $\BPD^\circ$ (green) on all graphs with respect to accuracy.

\section{Theoretical Properties of SBP} \label{sec:theory}
Here, we present some more formal arguments and discuss 
the properties of SBP to understand under which conditions the algorithm (presented in Section~\ref{sec:SBP}) can be expected to perform well.
We begin by formally defining the notion of a solution path in Section~\ref{subsec:theory:def} before we discuss the properties of SBP's solution path in Section~\ref{subsec:theory:properties}.
Finally, we restrict ourselves to unidirectional models, for which we analyze the accuracy of SBP in Section~\ref{subsec:theory:accuracy}.

\subsection{Solution Path: Definition}\label{subsec:theory:def}
First, we fix our notation: 
For the model $\PGM[m]$ with its potentials $\setOfPot_{m}$ we refer to the pseudomarginals by $\pseudomarginalsMinLocal(\scaling_m)$, and, with slight abuse of notation, we refer to the corresponding stationary point of the Bethe free energy by $\FBMinLocal(\scaling_m) = \FB(\pseudomarginalsMinLocal(\scaling_m))$.
It is beneficial to study the behavior of SBP as $M$ tends towards infinity. Therefore we consider the unit interval $\scaling \in [0,1]$ to be the compact support of the functions $\FB(\scaling)$ and $\pseudomarginals(\scaling)$.
SBP is inspired by the idea to proceed along a so-called \emph{solution path} as $\scaling$ increases from zero to one in order to obtain the marginal distributions for the model of interest.
Therefore, we consider a continuous homotopy function $H(\setOfMsg,\scaling): \mathbb{R}^{|\setOfMsg|+1} \rightarrow \mathbb{R}^{|\setOfMsg|}$.
Note that the scaling factor modifies the graphical model $\PGM = (\graph,\setOfPot(\scaling)$, for which the pairwise potentials are specified according to~\eqref{eq:scaling:potentials};
we define the homotopy accordingly by
\begin{align}
 H(\setOfMsg,\scaling) = \setOfMsg - \BP(\setOfMsg) \quad \text{where} \quad \setOfPot = \setOfPot(\scaling).\label{eq:homotopy}
\end{align}
Consequently, for a given value $\scaling$, $H(\setOfMsg,\scaling)$ is zero precisely for the fixed points $\setOfMsg[\circ]$ of $\PGM = (\graph,\setOfPot(\scaling)$.
It follows that a \emph{solution path} 
\begin{align}
 c(\scaling): H(\setOfMsg,\scaling) =0
\end{align}
 exists that (i) has a start point $c(\scaling=0) =\setOfMsg: H(\setOfMsg,\scaling=0) = 0$, (ii) an endpoint $c(\scaling=1) =\setOfMsg: H(\setOfMsg,\scaling=1) = 0$ , and (iii) is continuous over $\scaling \in [0,1]$, i.e., the solution path connects the start- with the endpoint.
 
SBP then proceeds along a solution path $c(\scaling)$ that implicitly defines the pseudomarginals $\pseudomarginalsMinLocal(\scaling)$ by~\eqref{eq:marginalsSingle}-\eqref{eq:marginalsPairwise}.
In particular, we refer to the start- and endpoint by $\pseudomarginalsMinLocal(\scaling=0)$ and $\pseudomarginalsMinLocal(\scaling=1)$ respectively.

\begin{figure}
 \centering
 \includegraphics[width=0.85\linewidth]{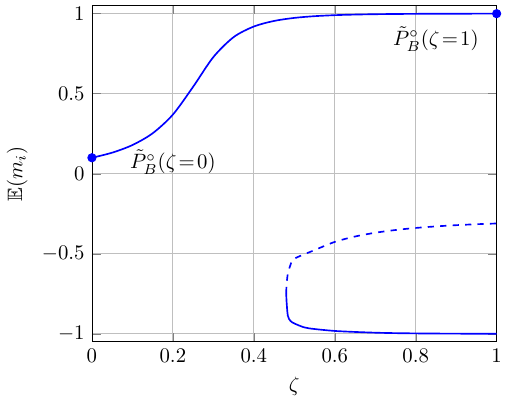}
 \caption{Solution path (as defined in Section~\ref{subsec:theory:def}) for Example~\ref{ex:solPath_nphc}. The depicted lines illustrate how the fixed points of BP evolve as $\scaling$ goes from zero to one. Solid lines correspond to minima of $\FB$; dashed line(s) to maxima of $\FB$. For more details see Example~\ref{ex:solPath_nphc}.}
 \label{fig:solution_path}
\end{figure}

The following example illustrates  the solution path for a unidirectional model.
\begin{ex}\label{ex:solPath_nphc}
Consider a unidirectional grid graph of size $3\times3$ with $\coupling{}{}=2$ and $\field{}=0.1$.
For this model, we can compute the fixed points of BP; i.e., the solutions to $ H(\setOfMsg,\scaling)=0$ with the numerical homotopy continuation method as in~\cite{knoll2017fixed}.

Fig.~\ref{fig:solution_path} shows how the solutions evolve as $\scaling$ goes from zero to one. 
Similar to Example~\ref{ex:unidirectional}, this illustrates how increasing the interactions between variables changes the energy landscape.
That is, for small values of $\scaling$ we have a unique fixed point until -- as $\scaling$ increases -- two additional fixed points emerge; one of which is a minimum and one a maximum of $\FB$.
Although multiple solutions to $ H(\setOfMsg,\scaling)=0$ exist for larger values of $\scaling$, all except for one lack a start point and are therefore of no relevance for any method that proceeds along a solution path defined by the homotopy in~\eqref{eq:homotopy}.
\end{ex}

\subsection{Solution Path: Properties}\label{subsec:theory:properties}
The following proposition summarizes the main properties of the solution path that is specified and followed by SBP.
\begin{prop}[Convergence Properties] 
SBP proceeds along a well-defined solution path $c(\scaling)$. 
More precisely, we have:
\begin{enumerate} 
 \item [(1)] BP has a unique fixed point $\setOfMsg[\circ]_1$ for $\scaling=0$, so that SBP has a unique start point $\pseudoTemplate{}{\stableOne}(\scaling=0)$
 (see Theorem~\ref{thm:init}).
 \item [(2)] A smooth (i.e., continuous) solution path originates from  $\pseudoTemplate{}{\stableOne}(\scaling=0)$
 (see Theorem~\ref{thm:smooth}).
 \item [(3)] SBP proceeds along this (unique) solution path $c(\scaling)$ and terminates after a fixed number of iterations (see Theorem~\ref{thm:complexity}).
\end{enumerate}
\label{prop:properties}
\end{prop}
In the sequel, we validate all points of Property~\ref{prop:properties}.

\subsubsection{Uniqueness of the Start Point}
First, we show that the initial model with $\scaling=0$ has a unique fixed point.
This is an immediate (and known) consequence of the fact that all variables are independent; we still include the following Theorem for the sake of completeness.
\begin{thm}[Property~\ref{prop:properties}.1]\label{thm:init}
For the initial model $\PGM[1]$ with parameters $\setOfPot(\scaling=0)$, BP has a unique fixed point $\setOfMsg[\circ]_1$.
That is, only a single start point exists for the solution path of SBP.
Moreover, this start point is exact, i.e.,  $\pseudomarginalsMinLocal(\scaling=0) =\pseudomarginalsExact(\scaling=0) $, and BP is guaranteed to converge.\footnote{ Note that this concurs with the sandwich-bound~\cite[Th.4]{weller2013bethe} that reduces to  
 $\marginals{\RV{i}}(+1) = {\field{i}}/(e^{\field{i}}+e^{-\field{i}}) = \marginalsApprox{\RV{i}}(+1)$ for $\coupling{i}{j}=0$. 
}
\end{thm}
\begin{IEEEproof}
First, we show that all messages converge to the identical value whenever all $\coupling{i}{j}=0$.
Therefore, note that we can omit the pairwise potential in the update equation of BP and rewrite~\eqref{eq:update} to
\begin{align}
 \msg[n+1]{i}{j}{} \propto\sum \limits_{x_i \in \stateSpace} \localShort{i} \!\!\! \prod \limits_{\RV{k} \in \neighbors{i} \backslash {j}} \!\!\! \msg[n]{k}{i}{}.
\label{eq:update_vanishing}
\end{align}
It follows that~\eqref{eq:update_vanishing} is independent of the state $x_j$, i.e., the computed messages are the same for all states of $X_j$.
Thus, after normalization, $\msg{i}{j}{} = \frac{1}{|\stateSpace|} = 0.5$ for all edges $\edge{i}{j} \in \setOfEdges$.

Second, the pseudomarginals are then obtained according to
\begin{align}
\marginalsShortApprox{x_i} &=\frac{1}{Z_i} \localShort{i}  \prod_{\RV{k} \in \neighbors{i}} \msg[\circ]{k}{i}{}\\
&= \frac{e^{x_i\field{i}}}{e^{\field{i}}+e^{-\field{i}}}.
\end{align}

Finally, we must show that the exact marginals are the same, i.e, that $\marginalsShort{x_i} = \marginalsShortApprox{x_i}$.
Therefore, note that the joint in~\eqref{eq:joint} factorizes only over the local potentials, which allows us to reorder the summation and compute the exact marginals according to
\begin{align}
 \marginalsShort{x_i} = \frac{1}{\PartitionGibbs} e^{x_i\field{i}} \hspace*{-0.5cm}\sum_{x_j:X_j\in\{\setOfNodes \backslash X_i\}} \prod_{X_j\in\{\setOfNodes \backslash X_i\}} \hspace*{-0.5cm} e^{x_j\field{j}} = \frac{e^{x_i\field{i}}}{e^{\field{i}}+e^{-\field{i}}}.
\end{align}
\end{IEEEproof}

Theorem~\ref{thm:init} thus reduces the problem of initializing SBP to computing the fixed point messages $\setOfMsg[\circ]_{1}$ of the initial model with $\scaling=0$; this can be done in linear time. 
Moreover, a unique start point implies that only a single solution path emerges from it.

\subsubsection{Smooth Solution Path}
Second, we show that -- for any model -- the Bethe free energy is a smooth function of the scaling parameter $\scaling$ in the sense that no discontinuities exist.
This implies the existence of a smooth solution path; 
let us make this notion of a smooth solution path more precise:

\begin{thm}[Property~\ref{prop:properties}.2]\label{thm:smooth}
Consider the pseudomarginals $\pseudomarginalsMinLocal(\scaling)$ along the (one) solution path that originates from the start point $\startp$.
Then, both the pseudomarginals $\pseudomarginalsMinLocal(\scaling)$ and the associated stationary point $\FBMinLocal(\scaling)$ are continuous functions of  the scaling parameter $\scaling$ along the whole solution path, i.e., for $\scaling \in [0,1]$.

\end{thm}
\begin{IEEEproof}
First, we show that the Bethe free energy $\FB(\pseudomarginals,\scaling)$ is continuously differentiable.
Therefore, the derivative with respect to $\scaling$ must exist and be continuous as well.
Note that $\FB(\pseudomarginals,\scaling)$ is a high-dimensional function defined over $\pseudomarginals\in \LP$, i.e., over all locally consistent marginals.
Now, let us consider~\eqref{eq:FB} with the pairwise potentials defined by~\eqref{eq:scaling:potentials}.
Then, when taking the derivative with respect to $\scaling$, all terms depending only on singleton marginals vanish and the derivative is given according to
  \begin{align}
  \frac{\partial \FB(\scaling)}{\partial \scaling} &=\!\! -\frac{\partial }{\partial \scaling}\!\!\sum_{\edge{i}{j} \in \setOfEdges} \sum_{x_i,x_j} \!\!\marginalsShortApprox{x_i,x_j} \ln \pairwiseShort{i}{j}  \nonumber  \\
    &= \!\!-\frac{\partial}{\partial \scaling}\!\!\sum_{\edge{i}{j} \in \setOfEdges} \sum_{x_i,x_j} \!\! \marginalsShortApprox{x_i,x_j} \cdot \scaling \coupling{i}{j} \cdot x_i x_j\nonumber\\
    & = - \sum_{\edge{i}{j} \in \setOfEdges} \coupling{i}{j} \cdot \correlation{i}{j}.  \label{eq:continuouslyDifferentiable}
  \end{align}
Since~\eqref{eq:continuouslyDifferentiable} is a finite sum over finite terms, this proves that $\FB(\scaling)$ is continuously differentiable.\footnote{ 
Moreover -- according to its definition in~\eqref{eq:FB} -- $\FB(\pseudomarginals,\scaling)$ is a polynomial, which guarantees that it is not just continuously differentiable but in fact an analytical function~\cite{rudin1964principles}. Note that this is in accordance with the fact that true phase transitions (singularities in the derivative of the free energy) can only occur in the thermodynamic limit, where \eqref{eq:continuouslyDifferentiable} becomes a sum over infinitely many terms that equates to infinity.} 
More precisely,~\eqref{eq:continuouslyDifferentiable} proves that $\FB(\pseudomarginals,\scaling)$ is continuously differentiable with respect to $\scaling$ for every point on the energy surface.

We specifically consider the local minimum $\FBMinLocal(\scaling)$ along the solution path, i.e., the minimum that emerges from  the unique start point $\FBMinLocal(\scaling=0)$.

Given that the Bethe free energy varies in a continuous fashion along this solution path (for $\scaling \in [0,1]$), the local minimum $\FBMinLocal(\scaling)$ and its position vary in a continuous fashion as well~\cite{dantzig}.\footnote{ 
This is actually not restricted to problems with a unique minimum; instead sets of minima are considered in~\cite{dantzig} to prove that minima of constrained minimization problems vary continuously under continuous changes of the objective function (see also~\cite[Section 5]{rockafellar}).}
For existence of this fixed point, note that minima of $\FB$ persist when increasing $\scaling$ (similar to increasing the inverse temperature; see~\cite{zdeborova2010generalization}). 
Consequently, because of the one-to-one correspondence between  stationary points of the Bethe free energy $\FBMinLocal(\scaling)$ and  fixed points of BP  $\pseudomarginalsMinLocal(\scaling)$, the pseudomarginals are also continuous.
Note, however, despite each minimum varying continuously with increasing $\scaling$, the global minimum  may not (as a former local minimum becomes the global minimum).

\end{IEEEproof}
In Theorem~\ref{thm:smooth}, we have shown that continuity of the Bethe free energy $\FB(\scaling)$ and the corresponding pseudomarginals $\pseudomarginalsMinLocal(\scaling)$ implies a smooth solution path.
We implicitly assumed that this further implies a well-behaved solution path.
In general this is the case but there are two scenarios that might be problematic as we will discuss now:

First, it is absolutely critical that the stationary points of $\FB$ are zero-dimensional, i.e., that every minimum is a distinct point, and that there are finitely many stationary points.
Fortunately, this is the case and the number of stationary points of $\FB$ is always finite~\cite{watanabe}; the same is true for the fixed points of BP (for as long as the messages are properly normalized)~\cite{martin2011}.

Second, a potential issue is how SBP copes with bifurcations of the solution path.
A bifurcation is a point along the solution path at which a minimum $\FBMinLocal$ turns into a maximum with two new minima branching off; 
i.e., one minimum splitting up into two new minima (see Example~\ref{ex:vanishing} for the illustration of a typical bifurcation).
In that case, SBP will follow the evolution of a single minimum and discard the other one.
Which fixed point is selected mainly depends on the implementation details of BP (this is because we rely on BP to find the next fixed point after our prediction). 
Note that different minima are usually wide apart (see Example~\ref{ex:vanishing} or~\cite{knoll_accuracy}).
If, however, two minima  would be close together, SBP might switch between branches because of numerical issues (e.g. because of quantization errors or a step-size too large). Since this is not a fundamental property of the solution path, we will assume sufficient numerical precision and rule out this phenomenon.
Thus, the existence of bifurcations is no particular problem and SBP retains the property of a unique solution path.

Although  bifurcations are  not problematic from a practical perspective, it is still relevant for our theoretical analysis of SBP to understand if and where bifurcations of the solution path exist.
Unfortunately, the current state of knowledge is still relatively limited in this respect.
Note, however, that at least for unidirectional models it is actually possible to make precise statements regarding the existence of bifurcations.

\begin{lm}\label{lm:bifurcation}
  Let $\graph$ be a unidirectional model. 
  Then, if the local potentials of all nodes are specified by $\field{i} = 0$, a bifurcation occurs. 
  Else, if some local potentials are nonzero, no bifurcation exists along the solution path.
\end{lm}
\begin{IEEEproof}
  Since we consider unidirectional models there are two minima of $\FB$ at most.
  We denote these minima by $\FB^{\stableOne}$ and $\FB^{\stableTwo}$.
  We will explain now, why these two minima branch off the previously unique minimum as we increase the scaling parameter $\scaling$ if and only if $\field{i} = 0$. 
  Therefore, we denote the critical scaling parameter, for which the two minima begin to exist, by $\scaling_c$.
  Then, by definition, a bifurcation exists if and only if both minima coincide for $\scaling_c$, i.e, if
  \begin{align}
  \lim_{\scaling \rightarrow \scaling_c^+} \mean{i}^{\stableOne} = \lim_{\scaling \rightarrow \scaling_c^+}\mean{i}^{\stableTwo}.
  \end{align}
  First, remember that both fixed points are symmetric if all $\field{i}=0$, i.e., $\mean{i}^{\stableOne} = -\mean{i}^{\stableTwo}$ (see Example~\ref{ex:vanishing} and Fig.~\ref{fig:energy:nolocal}).
  Because of this symmetry, both fixed points will bifurcate from a fixed point with zero mean $\mean{i}=0$.
  Note, that this specific case is well understood and serves as the archetype for the existence of bifurcations on the models we consider in this work~\cite{mooij2007sufficient}.
  
  Second, it remains to show that this is not the case if $\field{i} \neq 0$.
  Therefore, we show that both minima do not coincide for $\scaling_c$, i.e., 
  \begin{align}
  \lim_{\scaling \rightarrow \scaling_c^+} \mean{i}^{\stableOne} \neq \lim_{\scaling \rightarrow \scaling_c^+}\mean{i}^{\stableTwo}.
  \end{align}
  This is an immediate consequence of the fact that both fixed points are not symmetric anymore (see Section~\ref{subsec:solution_space}).
  In fact, $\mean{i}^{\stableOne}>0$ is strictly positive, whereas $\mean{i}^{\stableTwo}<0$ is strictly negative; 
  as a consequence, both fixed points can never coincide, which rules out the existence of bifurcations along the solution path.
\end{IEEEproof}
The implications of Lemma~\ref{lm:bifurcation} are illustrated in Example~\ref{ex:vanishing} and \ref{ex:unidirectional}.
Note how even small values for $\field{i}\neq 0$ break the symmetry between both fixed points and consequently imply the absence of bifurcations.

To summarize, Theorem~\ref{thm:smooth} and Lemma~\ref{lm:bifurcation} guarantee the existence of a smooth solution path that connects the start point $\pseudomarginalsMinLocal(\scaling=0)$ to the endpoint $\pseudomarginalsMinLocal(\scaling=1)$.
This is a property of fundamental importance as, at least in principle, it allows path-tracking algorithms (as SBP) to proceed along this well-behaved solution path.
The only remaining question is whether SBP is actually capable of proceeding along the solution path and if it obtains the endpoint; i.e., if SBP converges.

\subsubsection{Convergence Properties of SBP}

So far, we have substantiated the  claim that a smooth solution path emerges from the trivial start system leading to the target system.
The existence of this smooth solution path alone, however, is not sufficient to guarantee that SBP obtains the marginals of the target system.
As already discussed in Section~\ref{sec:SBP}, SBP relies on BP to proceed along the solution path; ideally, keeping the overall amount of BP iterations low.
If -- at some point (for $\scaling < 1$) -- the fixed point $\pseudomarginalsMinLocal(\scaling)$ becomes unstable, this approach will inevitably break down.
Nonetheless, SBP will proceed along the solution path for as long as it has a stable fixed point and approximates the marginals precisely at the onset of instability.
That is, even if the terminal fixed point $\pseudomarginalsMinLocal(\scaling=1)$ is unstable, SBP will always converge and approximate the marginals in a limited number of iterations.\footnote{ 
This is in stark contrast to plain BP that fails to converge and thus cannot be used to approximate the marginals in such a case.}
\begin{thm}[Property~\ref{prop:properties}.3]\label{thm:complexity}
 There exists some $\scaling \!\leq \!1$ so that SBP converges to $\tilde{P}_{B}^\circ(\scaling) \in \mathcal{L}$ in $\bigO(M N_{BP}\!)$.
\end{thm}
\begin{IEEEproof}
  SBP increases $\scaling$ as long as BP converges in less than $N_{BP}$ iterations, and stops otherwise. Consequently, BP   corrects the accuracy of the fixed point for 
  each value $\scaling_m$ within a bounded number of iterations. The run-time of SBP is further determined by the choice of $M$, i.e., the step-size (see Section~\ref{subsec:sbp:refinements}).
  That is, BP is run until convergence for at most $M$ times and thus, SBP converges in at most $M \cdot N_{BP}$ iterations.
\end{IEEEproof}

SBP is consequently capable of tracking the fixed point that emerges as $\scaling$ increases and requires $MN_{BP}$ iterations at most.
SBP may, however, only converge to a surrogate model for $\scaling_m < 1$ and is not guaranteed to obtain the pseudomarginals of the endpoint.
One can characterize this error by computing a bound on  $|\FBMinLocal(\scaling_m) - \FBMinLocal(\scaling=1)|$ given the difference between $\setOfPot(\scaling_m)$ and $\setOfPot(\scaling = 1)$ (cf.~\cite[Theorem 16]{ihler2005loopy}).

SBP works particularly well for attractive models, for which every minimum of $\FB$ is also a stable fixed point of BP~\cite{watanabe}.
More precisely, if for some scaling term $\scaling_m$ the fixed point of the solution path becomes unstable, i.e., if it turns into a local maximum, two stable fixed points will branch of~\cite[Theorem 6]{watanabe}.
Thus for attractive models (and also for balanced models) a smooth solution path with a stable fixed point always exits so that SBP is consequently capable of tracking the solution path until its end.

\begin{cor}
Let $\graph$ be an attractive or unidirectional model.
Then, every point along the solution path going from $\scaling=0$ to $\scaling=1$ is stable.
Consequently, some number of steps $M$ exists for which SBP converges and obtains the marginals $\pseudomarginalsMinLocal(\scaling=1)$.
\end{cor}
\begin{proof}
 According to Theorem~\ref{thm:smooth}, a smooth solution path $c(\scaling)$ exists that goes from the start point $\startp$ to the endpoint $\terminalp$, where every point along $c(\scaling)$ corresponds to a local minimum of the Bethe free energy $\FBMinLocal$.
 Every local minimum is also stable for attractive models~\cite{watanabe}.
 Therefore, the correction by BP will converge in every step if initialized sufficiently close to the fixed point, i.e., if the step size is chosen small enough. 
\end{proof}

\subsection{Accuracy of SBP}\label{subsec:theory:accuracy}
The convergence properties of SBP (Property~\ref{prop:properties}) are of fundamental importance for understanding SBP.
Yet, so far we have only been concerned if SBP converges but neglected any discussion regarding the quality of the obtained solution.
Thus, we will now theoretically analyze how well the fixed point obtained by SBP approximates the marginals.
Unfortunately, however, our understanding of what characterizes a fixed point that approximates the marginals well is relatively poor.
Indeed, it is not even obvious how the accuracy of the approximated marginals relates to the accuracy of the approximated free energy~\cite{knoll_accuracy}.

Therefore, we will mainly focus our discussion on the simplest model class for which the solution space is well understood (i.e., unidirectional models).
Note that for unidirectional models SBP is actually guaranteed to converge (see Section~\ref{subsec:solution_space}).
Regarding the accuracy of SBP -- when applied to unidirectional models -- we summarize the properties as follows:


\begin{prop}[Accuracy of SBP for Unidirectional Models]
Let $\graph$ be a unidirectional model, i.e., all edges are attractive ($\coupling{i}{j} > 0$) and all local potentials are specified by $\field{i}$ with the same sign. Without loss of generality we consider the case where all $\field{i}\geq0$.\footnote{
Note that identical results can be obtained in a straightforward manner for $\field{i}<0$ because of symmetry properties.}
Then, the solution path ends at a fixed point that approximates the marginals and the free energy well.
More specifically, SBP either obtains the exact marginals (if all $\field{i} = 0$) or finds the global minimum of the Bethe free energy $\FBMinGlobal$ that -- for unidirectional models -- corresponds to the fixed point with the most accurate marginals.
Let us now break down the argument into the following three points:
\begin{enumerate} 
 \item [(1)] For $\theta_i=0$, SBP obtains the exact marginals, i.e., $\pseudomarginalsMinLocal = \pseudomarginalsExact$ (see Theorem~\ref{thm:accuracy:fb_vanishing}).
 \item [(2)] For $\theta_i \neq 0$, SBP finds the global minimum of $\FB(\scaling)$ and gives the best approximation to the free energy $\FGibbs$ (see Theorem~\ref{thm:accuracy:fb}).
 \item [(3)] For $\theta_i \neq 0$, the global minimum of $\FB(\scaling)$ also gives the best approximation of the marginals, i.e., no other minimum of $\FB(\scaling)$ has more accurate marginals (see Theorem~\ref{thm:accuracy:marginals}).
\end{enumerate} 
\label{prop:attractive}
\end{prop}

First, we show that SBP obtains the exact marginals in the special case where all local potentials are specified by $\field{i}=0$
\begin{thm}[Property~\ref{prop:attractive}.1] \label{thm:accuracy:fb_vanishing}
 Consider a unidirectional model with $\field{i} = 0$.
 Then, the pseudomarginals obtained by SBP $\pseudoTemplate{}{\stableOne}(\scaling=1)$ are identical to the exact ones $\pseudomarginalsExact(\scaling=1)$.
\end{thm}
\begin{IEEEproof}
  For attractive models with $\field{i} = 0$ it is straightforward to compute the exact marginals.
  The exact marginals of all variables are uniformly distributed and have zero mean $\mean{i}(\scaling)=0$ for all values of $\scaling$;
  note that these marginals also correspond to a stationary point $\FBMinLocal(\scaling)$~\cite{mooij2007sufficient}.
  For sufficiently small values of $\coupling{i}{j}$ this stationary point is a (unique) minimum but for large values of $\coupling{i}{j}$ it turns into a maximum (see~\cite[pp.385]{mezard2009} and Example~\ref{ex:vanishing}).
  
  Even though the fixed point with $\mean{i}(\scaling)=0$ is unstable, SBP will still converge to this fixed point.
  This is rather surprising, as -- in general -- BP can neither converge to an unstable fixed point nor would it remain there because of quantization errors.
  There are two reasons why SBP obtains the potentially unstable fixed point in the special case of models with  $\field{i} = 0$ nonetheless.
  First, note that the (unique) fixed point of the initial mode is identical to the exact solution; hence all variables have zero mean at the start point, i.e., $\mean{i}^{\stableOne}(\scaling=0) = 0$.
  Second, all messages $\msg{i}{j}{} \in \setOfMsg[\circ](\scaling=0)$ take the same value of $\msg[\circ]{i}{j}{}= 1/2$, since $\field{i} = 0$.
  The extrapolated messages will stay the same for every iteration of SBP because -- as discussed above -- they already constitute a valid (albeit possibly unstable) fixed point (see Example~\ref{ex:vanishing}).
  Note, however, that these fixed point messages can be represented without quantization error in binary arithmetic.
  Thus SBP will remain on the fixed point corresponding to the exact marginals.
%
%
\end{IEEEproof}

The above statement arguably holds only for very restricted models and is thus of limited practical relevance.
Indeed, not only does standard BP obtain the same fixed point if initialized with identical messages but the exact marginals are also  known and can be computed analytically if all $\field{i} = 0$, obviating the need for approximate inference methods.

Our next results will discuss the accuracy of SBP for the more general class of unidirectional models both with respect to approximating the marginals and the Free energy (or partition function respectively).

Therefore, we first describe how the scaling term $\scaling$ affects the marginals along the solution path $c(\scaling)$ by generalizing Griffiths' inequality ~\cite{griffiths1967correlations} to the fixed points of BP.
In particular, this analysis reveals that the means $\meanMinLocal{i}$ are monotonically increasing with $\scaling$. 

\begin{lm}[Monotonicity of the Means]\label{lm:griffith}
 Consider two attractive models $\PGM[m]$ and $\PGM[n]$ on the same graph $\graph$, associated scaling parameters $\scaling_m < \scaling_n$, and local potentials specified by $\field{i}>0$.
 Note that we are effectively considering two nearly equivalent models with $\PGM[n]$ having strictly larger couplings than $\PGM[m]$.
 Now, let us consider a BP fixed point of $\PGM[m]$ with positive means $\meanMinLocal{i}(\scaling_m) > 0$.
 Then, the corresponding fixed point of $\PGM[n]$ has larger means $\meanMinLocal{i}(\scaling_n) > \meanMinLocal{i}(\scaling_m)$ and correlations  $\chi_{ij}^{\circ}(\scaling_n) \geq \chi_{ij}^{\circ}(\scaling_m)$.
\end{lm}
\begin{IEEEproof}
  The main ingredient of this proof is to show how the ratio between the messages for their respective states
  $\msgRatio{i}{j} = \frac{\msg{i}{j}{+1}}{\msg{i}{j}{-1}}$ increases monotonically with $\scaling$. 
  Then, the fact that the means and the correlations increase monotonically follows directly from the definition of the pseudomarginals in terms of the messages (see~\eqref{eq:marginalsSingle} and~\eqref{eq:marginalsPairwise}).

  First, we show that the message ratio $\msgRatio{i}{j}$ increases monotonically:
  therefore, let us denote the messages on both models by $\msg{(m)i}{j}{}$ and $\msg{(n)i}{j}{}$ respectively.
  Furthermore, without loss of generality, we assume that all couplings of $\PGM[n]$ are $\varepsilon$-larger than those of $\PGM[m]$.
  Then, for every $\varepsilon$ there exists $\delta>0$ so that whenever $0 < \msgRatio{(n)i}{j}-\msgRatio{(m)i}{j} < \delta$, we have $0<\coupling{(n)i}{j} -\coupling{(m)i}{j} < \varepsilon$.
  Note that by assumption $\mean{i} \in (0,1]$ so that 
  \begin{align}
  \msg{i}{j}{+1} \geq \msg{i}{j}{-1}. \label{eq:appendix:ratio}
  \end{align}
  First, we show that for all $\edge{i}{j} \in \setOfEdges$
  \begin{align}
  \frac{\msg[\circ]{(m)i}{j}{+1}}{\msg[\circ]{(m)i}{j}{-1}} < \frac{\msg[\circ]{(n)i}{j}{+1}}{\msg[\circ]{(n)i}{j}{-1}}.
  \label{eq:appendix:msgratio1}
  \end{align}
  Therefore, consider the update rules of~\eqref{eq:update} for both states 

  \begin{align}
  \msg[n+1]{(n)i}{j}{+1}&\propto   e^{\coupling{(m)i}{j} + \field{i} +\varepsilon} \!\!\!\!\!\!\!\!\! \prod \limits_{\RV{k} \in \neighbors{i} \backslash {j}} \!\!\!\!\!\!\!\!\! \msg[n]{(n)k}{i}{+1} \nonumber \\
  & \!\!\!\!\!\!\!\!\!\!\!\!\!\!\!\!\!\!  + e^{-\coupling{(m)i}{j} - \field{i} -\varepsilon} \!\!\!\!\!\!\!\!\! \prod \limits_{\RV{k} \in \neighbors{i} \backslash {j}} \!\!\!\!\!\!\!\!\! \msg[n]{(n)k}{i}{-1}, 
  \label{eq:appendix:update_pos}
  \end{align}
  \begin{align}
  \msg[n+1]{(n)i}{j}{-1}& \propto   e^{-\coupling{(m)i}{j} + \field{i} -\varepsilon} \!\!\!\!\!\!\!\!\! \prod \limits_{\RV{k} \in \neighbors{i} \backslash {j}} \!\!\!\!\!\!\!\!\! \msg[n]{(n)k}{i}{+1} \nonumber\\
  & \!\!\!\!\!\!\!\!\!\!\!\!\!\!\!\!\!\!+ e^{\coupling{(m)i}{j} - \field{i} +\varepsilon} \!\!\!\!\!\!\!\!\! \prod \limits_{\RV{k} \in \neighbors{i} \backslash {j}} \!\!\!\!\!\!\!\!\! \msg[n]{(n)k}{i}{-1}.
  \label{eq:appendix:update_neg}
  \end{align}
  In~\eqref{eq:appendix:update_pos} the larger product is multiplied by $e^{\varepsilon}$ and the smaller product is divided by $e^{\varepsilon}$. For~\eqref{eq:appendix:update_neg} it is exactly the other way round so that the ratio between the messages increases which proofs~\eqref{eq:appendix:msgratio1}. 
  We shall denote the imposed difference $\delta' \in \RealPositive$ on the messages by 
  \begin{align}
    \msg[\circ]{(n)i}{j}{+1} = \msg[\circ]{(m)i}{j}{+1}+\delta',\label{eq:appendix:msgdiff1}\\
    \msg[\circ]{(n)i}{j}{-1} = \msg[\circ]{(m)i}{j}{-1}-\delta'.\label{eq:appendix:msgdiff2}
  \end{align}
  Second, we show that $\meanMinLocal{i}(\scaling_n) > \meanMinLocal{i}(\scaling_m)$, which is an immediate consequence of plugging~\eqref{eq:appendix:msgdiff1} and~\eqref{eq:appendix:msgdiff2} into~\eqref{eq:marginalsSingle} (see Example~\ref{ex:vanishing} that illustrates how all fixed points, i.e., minima of $\FB$, are pulled towards $\meanMinLocal{i}$ extreme values of zero and one as $\coupling{}{}$ increases).
  
  Finally, it remains to show that $0 \stackrel{(i)}{\leq} \chi_{ij}^{\circ}(\scaling_m) \stackrel{(ii)}{\leq} \chi_{ij}^{\circ}(\scaling_n)$. Without loss of generality we assume that all variables have equal degree $d+1$, constant coupling strength $\coupling{i}{j}  = \coupling{}{}$, and constant local fields $\field{i} = \field{}$. 
  First we show that (i) holds, i.e., $\chi=\chi_{ij,0}^{\circ}$ is positive. Let us express the marginals by~\eqref{eq:marginalsPairwise} and denote the messages by $\mu = \msg{(m)i}{j}{1}$. It follows that $\msg{(m)i}{j}{-1} = (1-\mu)$ and that
  \begin{align}
    \chi\! &=\! e^{\coupling{}{}+2\field{}} \mu^{2d}\!\! + \!e^{\coupling{}{}-2\field{}} (1\!-\!\mu)^{2d}\!\! - 2 e^{-J} \mu^{d} (1\!-\!\mu)^{d}.
    \label{eq:correlation}
  \end{align}
  Let us further represent the messages by $\mu = 1/2+\beta$ with $\beta \in [0,1/2]$. It follows that
  \begin{align}
    \chi \stackrel{(a)}{\geq} &\left(\left(1/2+\beta\right)^{2d} \! + \!\left(1/2-\beta\right)^{2d}\right) \!-\! 2 \left(1/2+\beta\right)^d\left(1/2-\beta\right)^d \nonumber \\
    = &\left(\left(1/2-\beta\right)^d-\left(1/2+\beta\right)^d\right)^2 \label{eq:correlation:positive:3}\\
    \stackrel{(b)}{\geq}  & 0,  
  \end{align}
  where $(a)$ follows from neglecting all exponential terms and thus upper bounding the positive term and lower bounding the negative term (with equality if and only if $\coupling{}{} = 0$ and $\field{} = 0$) and
  $(b)$ is a direct consequence of the square in~\eqref{eq:correlation:positive:3}.
  Now let us show that (ii) holds, i.e., 
  $\chi$ increases monotonically, by taking the derivative of~\eqref{eq:correlation}, so that
  \begin{align}
    \frac{\partial}{\partial \mu}\chi = 
    & 2d \left(e^{\coupling{}{}+2\field{}} \mu^{2d-1} - e^{\coupling{}{}-2\field{}} \left(1-\mu\right)^{2d-1}\right) \nonumber\\
    &+\!\! 2 d e^{-J}\!\! \left( \mu^{d} (1-\mu)^{d-1} \!\!- \mu^{d-1} (1-\mu)^{d} \right) \label{eq:correlation:monotone:1}\!\\
  \stackrel{(a)}{\geq} &  2 d e^{-J} \left( \mu^{d} (1-\mu)^{d-1} - \mu^{d-1} (1-\mu)^{d} \right) \label{eq:correlation:monotone:2}\\
  \stackrel{(b)}{\geq} & 0,
    \end{align}
  where $(a)$ follows from neglecting the, strictly positive, first term in~\eqref{eq:correlation:monotone:1}, and $(b)$ 
  is a direct consequence from~\eqref{eq:appendix:ratio}.
\end{IEEEproof}


\begin{thm}[Property~\ref{prop:attractive}]\label{thm:accuracy:fb}
Consider a unidirectional model, i.e, an attractive model with $\field{i} > 0$.
Then the minimum of the Bethe free energy along the solution path $\FBMinLocalSpecific{\stableOne}(\scaling)$ is always the global minimum, i.e., $\FBMinLocalSpecific{\stableOne}(\scaling) = \min \FB(\scaling)$.
Moreover, the Bethe free energy of this stationary point decreases monotonically with $\scaling$.

\end{thm}

\begin{IEEEproof}
  A unique start point $\pseudoTemplate{}{\stableOne}(\scaling=0)$ exists by Theorem~\ref{thm:init}.
  For $\field{i}>0$ this fixed point has only positive means $\meanMinLocalSpecific{i}{\stableOne}(\scaling=0)>0$ and correlations $\correlationMinLocalSpecific{i}{j}{\stableOne}(\scaling=0)>0$.
  
  Consequently, Lemma~\ref{lm:griffith} applies, which implies that the corresponding means $\mean{i}^{\stableOne}(\scaling)$ and correlations $\chi_{ij}^{\stableOne}(\scaling) $ are monotonically increasing along the solution path. 
  This further implies that the Bethe free energy $\FBMinLocal(\scaling)$ decreases along the solution path.
  Let $\FB^{\stableOne}(\scaling=1)$ correspond to the endpoint of the solution path $c(\scaling)$ that 
  emerges from the origin; then, it immediately follows that the error with respect to the endpoint $\FB^{\stableOne}(\scaling=1)$ decreases along the solution path: i.e., consider two arbitrary values $m,n \in [0,1]$ such that $n>m$, then   
  $|\FB^{\stableOne}(\scaling_m)-\FB^{\stableOne}(\scaling=1)| \geq |\FB^{\stableOne}(\scaling_{n})-\FB^{\stableOne}(\scaling = 1)|$.

  It remains to show that SBP obtains the fixed point $\FB^{\stableOne}(\scaling)$ that is the global minimum of the Bethe free energy for a given scaling parameter $\scaling$.
  Therefore, consider the fact that unidirectional models have two fixed points at most, only one of which has positive means (see Lemma~\ref{lm:symmetry:not}).
  As discussed above, SBP proceeds along a solution path belonging to the consistent fixed point with $\mean{i}^{\stableOne}(\scaling)>0$.
  The second fixed point $\stableTwo$, if it exists, has negative means $\mean{i}^{\stableTwo}(\scaling) <0$.
  Given that SBP obtains the consistent fixed point, it follows from Lemma~\ref{lm:symmetry:not} that SBP obtains the global minimum, i.e., $\FB^{\stableTwo} \geq \FB^{\stableOne} = \FBMinGlobal$.
\end{IEEEproof}

Theorem~\ref{thm:accuracy:fb} makes it explicit that -- for unidirectional models -- SBP favors the fixed point that defines the global minimum of $\FB$. 
This is in accordance with the observation that the average energy is linear in the pseudomarginals (see~\eqref{eq:energy}), which further implies that varying the local potentials essentially tilts the energy landscape (cf.~\cite{pitkow2011learning}).
Note that this behavior is also illustrated in Example~\ref{ex:vanishing} and~\ref{ex:unidirectional} accordingly:
i.e, as opposed to Fig.~\ref{fig:energy:nolocal}, local potentials with $\field{i} > 0$ tilt the energy landscape in Fig.~\ref{fig:energy:local} so that the consistent fixed point constitutes the global minimum of $\FB$.

Although finding the global minimum of $\FB$ is desirable, one is ultimately interested in approximating the exact free energy as good as possible.
Fortunately, both properties coincide for unidirectional models.

\begin{cor}[Accuracy of the Bethe Approximation]
Consider a unidirectional model.
Then there is no other fixed point of BP $\pseudoTemplate{}{k}$ that approximates the free energy $\FGibbs$ better than the one obtained by SBP, i.e., 
\begin{align}
\pseudoTemplate{}{\stableOne}(\scaling) =\argmin\limits_{\pseudoTemplate{}{k}(\scaling)\in\LP}\! |\FB^{k}(\scaling)\! -\! \FGibbs^*\!|. \nonumber
\end{align}
\end{cor}
\begin{IEEEproof}
This is a direct consequence of SBP obtaining the global minimum of the Bethe free energy (see Theorem~\ref{thm:accuracy:fb}) and the fact that the Bethe free energy upper bounds the free energy for attractive models (cf.~\cite{ruozzi2013bethebound}).
 \end{IEEEproof}
This implies that the fixed point obtained by SBP minimizes the approximation error of the free energy $|\FB^{k}(\scaling)\! -\! \FGibbs^*\!|$. 
Moreover -- because of the direct relationship between the Bethe free energy and the Bethe approximation of the partition function -- this implies that the fixed point obtained by SBP also approximates the partition function best.

Note, however, that statements about the approximation quality of the free energy do not necessarily extend to the accuracy of the pseudomarginals~\cite{knoll_accuracy}.
Nonetheless, at least for unidirectional models, this is the case and SBP obtains the fixed point with the most accurate marginals.


\begin{thm}[Accuracy of Marginals]\label{thm:accuracy:marginals}
Consider a unidirectional model and assume that the RSB assumption holds (see Conjecture~\ref{lm:rsb}).
Then there is no other fixed point of BP $\pseudoTemplate{}{k}$ that approximates the marginals better than the one obtained by SBP, i.e, 
\begin{align}
\pseudoTemplate{}{\stableOne}(\scaling) = \argmin\limits_{\pseudoTemplate{}{k}(\scaling)\in\LP}\! |\pseudoTemplate{}{k}(\scaling) - \pseudomarginalsExact(\scaling)|. \nonumber
\end{align}
\end{thm}
\begin{IEEEproof}
Note that we are only considering unidirectional models here.
For these models, although not formally verified, the RSBP assumption is accepted to hold.
Thus, if multiple fixed points do exist this allows us to express the exact marginals as a convex combination of all fixed points according to Conjecture~\ref{lm:rsb}.

Moreover, remember that the Bethe free energy of unidirectional models has two minima at most (see Section~\ref{subsec:solution_space}).
This -- and the fact that the convex combination of both fixed points yields the exact marginals by assumption -- implies that the fixed point obtained by SBP (i.e., the one that minimizes the Bethe free energy) approximates the marginals best.
\end{IEEEproof}

To conclude, the fixed point along the solution path is always stable for unidirectional models.
Thus, SBP follows the solution path to its endpoint.
In doing so, SBP not only decreases the Bethe free energy and obtains the global minimum but it also provides the fixed point with the most accurate marginals.

\section{Conclusion}\label{sec:conclusion}
In this paper, we introduced an iterative algorithm to perform approximate inference: self-guided belief propagation (SBP) is a simple and robust method that gradually accounts for the pairwise potentials and guides itself towards a unique, stable, and accurate solution.

We provide a comprehensive theoretical analysis in order to validate the underlying assumptions:
in particular, we showed that: 
(i) a smooth solution path exists and originates from a unique start point that is obtained by neglecting the pairwise potentials;
(ii) this solution path is well-behaved and can be tracked by SBP; and
(iii) for \emph{unidirectional models}, the solution of SBP approximates the exact solution well and corresponds to the global optimum of the Bethe approximation.

The theoretical analysis does not fully answer for which models SBP is expected to perform particularly well.
Although SBP obtains the most accurate fixed point for unidirectional models, the marginals will not be approximated well for models with strong couplings, for the simple reason that accurate fixed points do not exist at all.
We expect that SBP is particularly advantageous for all models that have a large amount of fixed points with varying accuracy.
Thus SBP will perform well for dense models with random couplings (particularly if they are centered around zero) for which the marginals are largely affected by the local potentials.
Not only do we expect that SBP approximates the marginals well in that case but it is actually essential for estimating marginals as BP will often fail to converge.

In order to verify SBP empirically, we applied SBP to a range of models, not only attractive but also general ones.
In our experiments, SBP consistently improves the accuracy in comparison with BP (with and without damping). 
Besides, SBP approximates the exact marginals well on graphical models for which BP does not converge at all.  
\def\bibfont{\small}

\section*{Acknowledgment}
The authors greatly appreciate the valuable discussions with Bernhard C. Geiger, KNOW-Center, and acknowledge his help in making the formal definition of SBP much more concise.
Moreover, we thank Alexander Ihler, University of California Irvine, for the encouraging discussion during UAI'17 that was pivotal for actually pursuing the -- at this time -- rather unpolished idea.
We further appreciate the help of Florian Kulmer in implementing and empirically evaluating our algorithm and are particularly indebted to Harald Leisenberger for carefully reading the manuscript.

This work was supported by the Graz University of Technology LEAD project ``Dependable Internet of Things in Adverse Environments''.
Adrian Weller acknowledges support from the David MacKay Newton research fellowship at Darwin College, The Alan Turing Institute under EPSRC grant EP/N510129/1 \& U/B/000074, and the Leverhulme Trust via the Leverhulme Centre for the Future of Intelligence (CFI).

\bibliographystyle{abbrv}
\bibliography{SBP}

%

\begin{IEEEbiography}[{\includegraphics[width=1in,height=1.25in,clip,keepaspectratio]{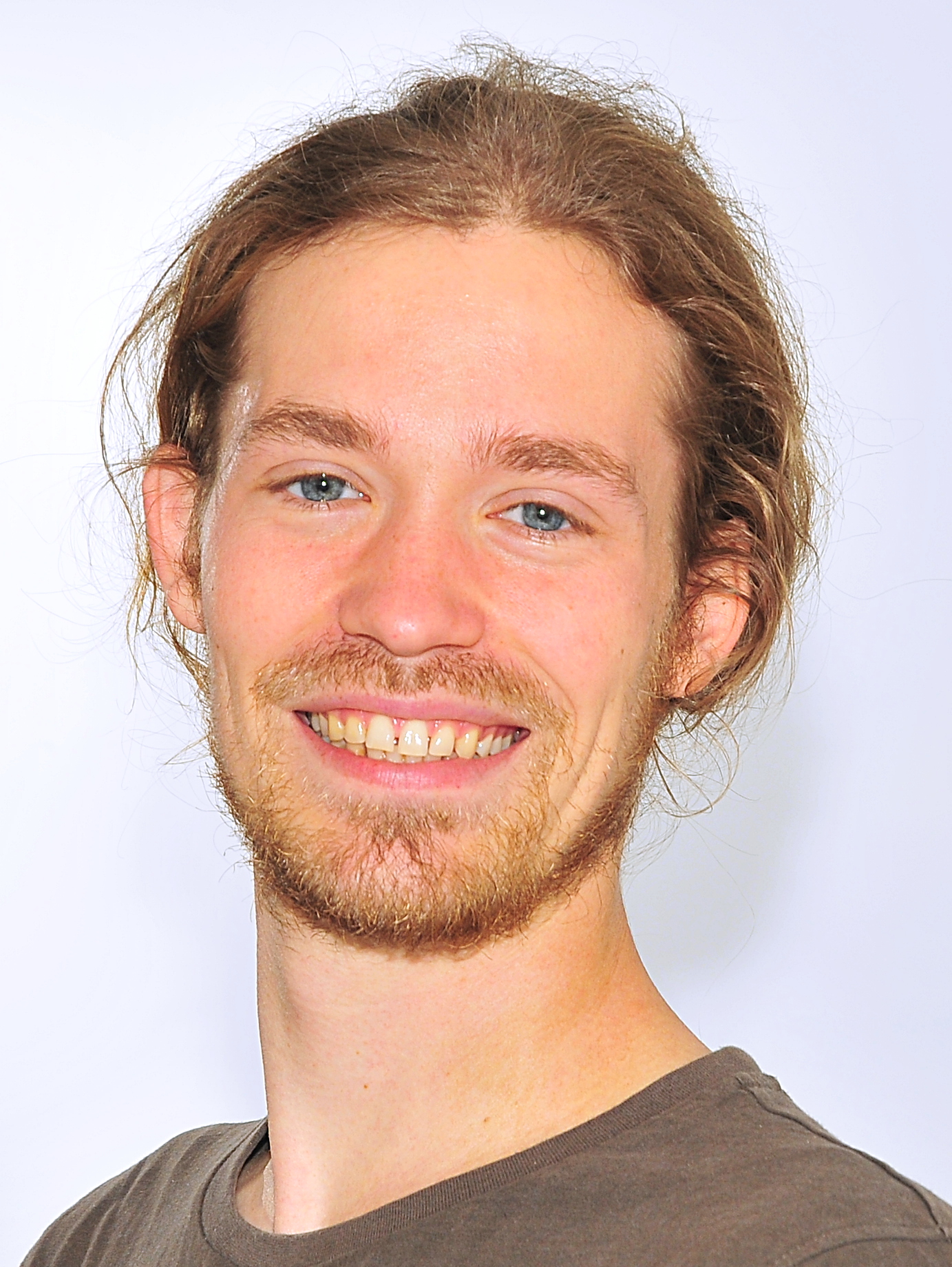}}]{Christian Knoll}
received his Bsc and Msc degree in Information and Computer Engineering in 2011 and 2014 and earned his PhD degree in 2019 from Graz University of Technology. 
He is currently a postdoctoral researcher with the Signal Processing and Speech Communication Laboratory at Graz University of Technology.
His research interests include machine learning, graphical models, and statistical signal processing with a particular focus on message passing methods for probabilistic inference.
\end{IEEEbiography}

\begin{IEEEbiography}[{\includegraphics[width=1in,height=1.25in,clip,keepaspectratio]{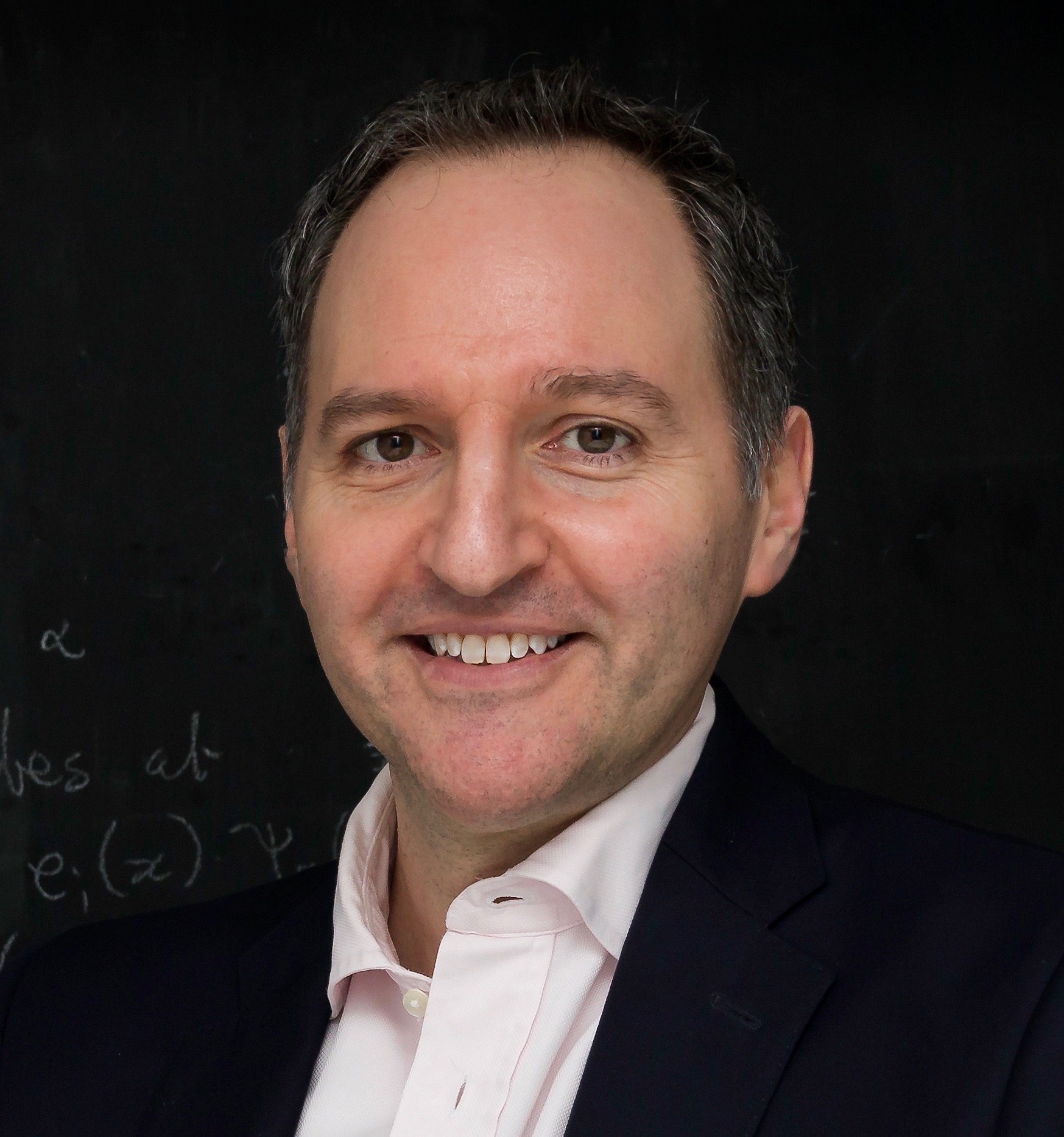}}]{Adrian Weller} received an MA in mathematics from the University of Cambridge, and a PhD in Computer Science from Columbia University. He 
is Programme Director for AI at The Alan Turing Institute, the UK national institute for data science and AI, where he is also a Turing Fellow leading work on safe and ethical AI. He is a Principal Research Fellow in Machine Learning at 
Cambridge, and at the Leverhulme Centre for the Future of Intelligence where he is Programme Director for Trust and Society. His interests span AI, its commercial applications and helping to ensure beneficial outcomes for society. 
\end{IEEEbiography}

\begin{IEEEbiography}[{\includegraphics[width=1in,height=1.25in,clip,keepaspectratio]{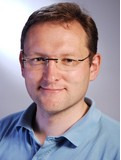}}]{Franz Pernkopf}
received his MSc (Dipl. Ing.) degree in Electrical
Engineering at Graz University of Technology, and a PhD degree
from the University of Leoben. In 2002 he was awarded the Erwin
Schr\"{o}dinger Fellowship. He was a Research Associate at the University of Washington,
Seattle, from 2004 to 2006. From 2010-2019 he was an Associate Professor at
the Laboratory of Signal Processing and Speech Communication; since 2019, he is a Professor for
Intelligent Systems, at Graz University of Technology. His
research is focused on pattern recognition, machine
learning, and computational data analytics with applications in various
fields ranging 
from signal processing to medical data analysis and industrial applications. 

\end{IEEEbiography}

\end{document}

%% file: SBP_pseudocode.tex
\label{appendix:A}
Pseudocode of SBP is presented in Algorithm~\ref{alg:sbp}. 
Note that we initialize the messages $\setOfMsg[\init]$ randomly.
The sequence of messages over all models $\{\PGM[m] \}$ is collected in $\{\setOfMsg[\circ]_\iteration{m}\} = \{\setOfMsg[\circ]_\iteration{1},\ldots,\setOfMsg[\circ]_\iteration{m}\}$.
\textsf{ExtrapolateMsg} applies cubic spline extrapolation to estimate the initial messages of the subsequent model.

We further present the pseudocode for the adaptive step size controller in Algorithm~\ref{alg:adaptivestepsize}.
The step size is increased depending on $k$, where $k$ is the number of preceeding fixed point messages that are $\varepsilon$-close to the current fixed point messages $\setOfMsg[\circ]_\iteration{m}$

\begin{algorithm}
\caption{Self-Guided Belief Propagation (SBP)}\label{alg:sbp}
\LinesNumbered
\DontPrintSemicolon
\SetKwFunction{BP}{BP}\SetKwFunction{Add}{add}\SetKwFunction{Extrapolate}{ExtrapolateMsg}\SetKwFunction{ScalePotentials}{ScalePotentials}\SetKwFunction{AdaptiveStepsize}{AdaptiveStepSize}
\SetKwInOut{Input}{input}\SetKwInOut{Output}{output}
\Indm
  \Input{graph $\mathcal{G} = (\mathbf{X},\mathbf{E})$, potentials $\setOfPot$,\\
  initial step-size $step_{init}$, \\
  maximum number of iterations $N_{BP}$}
  \Output{fixed point messages $  \setOfMsg[\circ] $}
  \BlankLine
\Indp
  initialization $ \setOfMsg[\init]_\iteration{1} \leftarrow \setOfMsg[\init] $\;
  \BlankLine
  $m \leftarrow 1$\;
  $\scaling_1\leftarrow 0$\;
  \While{$\scaling_m \le 1$}
   {
    $\setOfPot(\scaling_m)$ $\leftarrow$ \ScalePotentials{$\setOfPot,\scaling_m$}\;
    $(\setOfMsg,iterations)$ $\leftarrow$ \BP{$\setOfMsg[\init]_\iteration{m},\setOfPot(\scaling_m),N_{BP}$}\;
    \eIf{ $iterations < N_{BP}$}
    {   $\setOfMsg[\circ]_\iteration{m} \leftarrow \setOfMsg$\; 
    }
    { break\;	    
    }

    \eIf{ $ adaptive \ stepsize $}
	{   
	$\scaling_{m+1} \leftarrow \scaling_m +$ \AdaptiveStepsize{$ \{\setOfMsg[\circ]_\iteration{m}\} $, $ step_{init} $, $ m $}\;
	}
	{     $\scaling_{m+1} \leftarrow \scaling_m + step_{init}$\; 
	}

    $\setOfMsg[\init]_\iteration{m+1}$ $\leftarrow$ \Extrapolate{$\{\setOfMsg[\circ]_\iteration{m}\}$,$\{\scaling_m\}$}\;
    $m \leftarrow m+1$\;
   }
  $ \setOfMsg[\circ] \leftarrow \setOfMsg[\circ]_\iteration{m-1}  $
\end{algorithm}
%

\begin{algorithm}
	\caption{Adaptive Step Size Controller}\label{alg:adaptivestepsize}
	\LinesNumbered
	\DontPrintSemicolon
	\SetKwFunction{Add}{add}\SetKwFunction{ComputeMarginals}{ComputeMarginals}\SetKwFunction{UnnormalizedProbability}{UnnormalizedMarginals}\SetKwFunction{MSE}{MSE}
	\SetKwInOut{Input}{input}\SetKwInOut{Output}{output}
	\Indm
	\Input{sequence of messages $ \{\setOfMsg[\circ]_\iteration{m}\} $, $ step_{init} $, $ m $\\ threshold $\varepsilon$}
	\Output{stepsize $ step $}
	\BlankLine
	\Indp
	$step \leftarrow step_{init}$\;
	\BlankLine
	$k \leftarrow 1$\;
	\While{  $ |\setOfMsg[\circ]_\iteration{m} - \setOfMsg[\circ]_\iteration{m-k} |^2  < \varepsilon $}
	{
		$k \leftarrow k + 1$\;
		$step \leftarrow step + step_{init} \cdot k$\;
	}
\end{algorithm}
\renewcommand{\BP}{\mathcal{BP}}

%% file: SBP.bbl
\begin{thebibliography}{10}

\bibitem{allgower2003numerical}
E.~Allgower and K.~Georg.
\newblock {\em {Introduction to Numerical Continuation Methods}}.
\newblock Society for Industrial and Applied Mathematics, 2003.

\bibitem{belanger1991random}
D.~Belanger and A.~Young.
\newblock The random field {Ising} model.
\newblock {\em Journal of magnetism and magnetic materials}, 100(1-3):272--291,
  1991.

\bibitem{braunstein2007encoding}
A.~Braunstein, F.~Kayhan, G.~Montorsi, and R.~Zecchina.
\newblock Encoding for the {B}lackwell channel with reinforced belief
  propagation.
\newblock In {\em IEEE International Symposium on Information Theory}, 2007.

\bibitem{braunstein2005survey}
A.~Braunstein, M.~M{\'e}zard, and R.~Zecchina.
\newblock {Survey propagation: an algorithm for satisfiability}.
\newblock {\em Random Structures \& Algorithms}, 27(2):201--226, 2005.

\bibitem{chandrasekaran2011counting}
V.~Chandrasekaran, M.~Chertkov, D.~Gamarnik, D.~Shah, and J.~Shin.
\newblock {Counting independent sets using the {Bethe} approximation}.
\newblock {\em SIAM Journal on Discrete Mathematics}, 25(2):1012--1034, 2011.

\bibitem{coja2019bethe}
A.~Coja-Oghlan and W.~Perkins.
\newblock Bethe states of random factor graphs.
\newblock {\em Communications in Mathematical Physics}, 366(1):173--201, 2019.

\bibitem{cooper1990computational}
G.~F. Cooper.
\newblock {The computational complexity of probabilistic inference using
  Bayesian belief networks}.
\newblock {\em Artificial intelligence}, 42(2-3):393--405, 1990.

\bibitem{dantzig}
G.~B. Dantzig, J.~Folkman, and N.~Shapiro.
\newblock On the continuity of the minimum set of a continuous function.
\newblock {\em Journal of Mathematical Analysis and Applications}, 17:519--548,
  1967.

\bibitem{dembo2010ising}
A.~Dembo and A.~Montanari.
\newblock {Ising models on locally tree-like graphs}.
\newblock {\em The Annals of Applied Probability}, 20(2):565--592, 2010.

\bibitem{elidan2012residual}
G.~Elidan, I.~McGraw, and D.~Koller.
\newblock {Residual belief propagation: Informed scheduling for asynchronous
  message passing}.
\newblock In {\em {Proceedings of {UAI}}}, 2006.

\bibitem{globerson2007convergent}
A.~Globerson and T.~Jaakkola.
\newblock Convergent propagation algorithms via oriented trees.
\newblock In {\em Proceedings of {UAI}}, 2007.

\bibitem{goldberg2003computational}
L.~A. Goldberg, M.~Jerrum, and M.~Paterson.
\newblock The computational complexity of two-state spin systems.
\newblock {\em Random Structures \& Algorithms}, 23(2):133--154, 2003.

\bibitem{griffiths1967correlations}
R.~B. Griffiths.
\newblock Correlations in {Ising} ferromagnets. ii. external magnetic fields.
\newblock {\em Journal of Mathematical Physics}, 8(3):484--489, 1967.

\bibitem{harary1953}
F.~Harary et~al.
\newblock On the notion of balance of a signed graph.
\newblock {\em The Michigan Mathematical Journal}, 2(2):143--146, 1953.

\bibitem{hazan2008convergent}
T.~Hazan and A.~Shashua.
\newblock {Convergent message-passing algorithms for inference over general
  graphs with convex free energies}.
\newblock In {\em {Proceedings of {UAI}}}, 2008.

\bibitem{ihler2005loopy}
A.~Ihler, J.~Fisher, and A.~Willsky.
\newblock {Loopy belief propagation: convergence and effects of message
  errors}.
\newblock In {\em {Journal of Machine Learning Research}}, pages 905--936,
  2005.

\bibitem{jerrum1993polynomial}
M.~Jerrum and A.~Sinclair.
\newblock Polynomial-time approximation algorithms for the {Ising} model.
\newblock {\em SIAM Journal on Computing}, 22(5):1087--1116, 1993.

\bibitem{johnson2016learning}
J.~K. Johnson, D.~Oyen, M.~Chertkov, and P.~Netrapalli.
\newblock Learning planar {Ising} models.
\newblock {\em The Journal of Machine Learning Research}, 17(1):7539--7564,
  2016.

\bibitem{jordan2004}
M.~Jordan.
\newblock {Graphical models}.
\newblock {\em Statistical Science}, pages 140--155, 2004.

\bibitem{knoll2017fixed}
C.~Knoll, D.~Mehta, T.~Chen, and F.~Pernkopf.
\newblock {Fixed Points of Belief Propagation--An Analysis via Polynomial
  Homotopy Continuation}.
\newblock {\em IEEE Trans. on Pattern Analysis and Machine Intelligence}, 2017.

\bibitem{knoll2017uai}
C.~Knoll and F.~Pernkopf.
\newblock {On Loopy Belief Propagation--Local Stability Analysis for
  Non-Vanishing Fields}.
\newblock In {\em {Proceedings of {UAI}}}, 2017.

\bibitem{knoll_accuracy}
C.~Knoll and F.~Pernkopf.
\newblock Belief propagation: Accurate marginals or accurate partition function
  -- where is the difference?
\newblock In {\em Proceedings of {UAI}}, 2019.

\bibitem{knoll2015message}
C.~Knoll, M.~Rath, S.~Tschiatschek, and F.~Pernkopf.
\newblock {Message Scheduling Methods for Belief Propagation}.
\newblock In {\em {Machine Learning and Knowledge Discovery in Databases}},
  pages 295--310. Springer, 2015.

\bibitem{koehler2019fast}
F.~Koehler.
\newblock Fast convergence of belief propagation to global optima: Beyond
  correlation decay.
\newblock In {\em Proceedings of {NeurIPS}}, 2019.

\bibitem{koller-friedman}
D.~Koller and N.~Friedman.
\newblock {\em {Probabilistic Graphical Models: Principles and Techniques}}.
\newblock MIT press, 2009.

\bibitem{kschischang2001factor}
F.~Kschischang, B.~Frey, and H.~Loeliger.
\newblock {Factor graphs and the sum-product algorithm}.
\newblock {\em IEEE Trans. on Information Theory}, 47(2), 2001.

\bibitem{martin2011}
V.~Martin, J.-M. Lasgouttes, and C.~Furtlehner.
\newblock The role of normalization in the belief propagation algorithm.
\newblock {\em preprint arXiv:1101.4170}, 2011.

\bibitem{meltzer2009convergent}
T.~Meltzer, A.~Globerson, and Y.~Weiss.
\newblock {Convergent message passing algorithms: a unifying view}.
\newblock In {\em Proceedings of {UAI}}, 2009.

\bibitem{meshi2009convexifying}
O.~Meshi, A.~Jaimovich, A.~Globerson, and N.~Friedman.
\newblock {Convexifying the {Bethe} free energy}.
\newblock In {\em {Proceedings of {UAI}}}, 2009.

\bibitem{mezard2009}
M.~Mezard and A.~Montanari.
\newblock {\em {Information, Physics, and Computation}}.
\newblock Oxford Univ. Press, 2009.

\bibitem{mezard1987spin}
M.~Mezard, G.~Parisi, and M.~Virasoro.
\newblock {\em {Spin Glass Theory and Beyond: An Introduction to the Replica
  Method and Its Applications}}, volume~9.
\newblock World Scientific Publishing Co Inc, 1987.

\bibitem{mooij2005properties}
J.~Mooij and H.~Kappen.
\newblock {On the properties of the {Bethe} approximation and loopy belief
  propagation on binary networks}.
\newblock {\em Journal of Statistical Mechanics: Theory and Experiment},
  2005(11):P11012, 2005.

\bibitem{mooij2007sufficient}
J.~M. Mooij and H.~J. Kappen.
\newblock {Sufficient conditions for convergence of the sum--product
  algorithm}.
\newblock {\em IEEE Trans. on Information Theory}, 53(12):4422--4437, 2007.

\bibitem{murphy1999loopy}
K.~Murphy, Y.~Weiss, and M.~Jordan.
\newblock {Loopy belief propagation for approximate inference: an empirical
  study}.
\newblock In {\em {Proceedings of {UAI}}}, 1999.

\bibitem{pernkopf2014pgm}
F.~Pernkopf, R.~Peharz, and S.~Tschiatschek.
\newblock {\em {Introduction to Probabilistic Graphical Models}}.
\newblock Academic Press{\rq} Library in Signal Processing, 2014.

\bibitem{perugini}
G.~Perugini and F.~Ricci-Tersenghi.
\newblock Improved belief propagation algorithm finds many {Beth}e states in
  the random-field {Ising} model on random graphs.
\newblock {\em Physical Review E}, 97(1):012152, 2018.

\bibitem{pitkow2011learning}
X.~Pitkow, Y.~Ahmadian, and K.~D. Miller.
\newblock Learning unbelievable probabilities.
\newblock In {\em Proceedings of {NIPS}}, 2011.

\bibitem{rockafellar}
R.~T. Rockafellar and R.~J.-B. Wets.
\newblock {\em Variational Analysis}, volume 317.
\newblock Springer Science \& Business Media, 2009.

\bibitem{rudin1964principles}
W.~Rudin.
\newblock {\em Principles of mathematical analysis}, volume~3.
\newblock McGraw-hill New York, 1964.

\bibitem{ruozzi2013bethebound}
N.~Ruozzi.
\newblock {Beyond log-supermodularity: lower bounds and the {Bethe} partition
  function}.
\newblock In {\em {Proceedings of {UAI}}}, 2013.

\bibitem{saade2017spectral}
A.~Saade, F.~Krzakala, and L.~Zdeborov{\'a}.
\newblock Spectral bounds for the {Ising} ferromagnet on an arbitrary given
  graph.
\newblock {\em Journal of Statistical Mechanics: Theory and Experiment},
  2017(5):053403, 2017.

\bibitem{shin2012complexity}
J.~Shin.
\newblock {Complexity of {Bethe} Approximation.}
\newblock In {\em {Proceedings of {AISTATS}}}, pages 1037--1045, 2012.

\bibitem{sommese2005numerical}
A.~Sommese and C.~Wampler.
\newblock {\em {The Numerical Solution of Systems of Polynomials Arising in
  Engineering and Science}}, volume~99.
\newblock World Scientific, 2005.

\bibitem{sontag2008new}
D.~Sontag and T.~S. Jaakkola.
\newblock New outer bounds on the marginal polytope.
\newblock In {\em Proceedings of {NIPS}}, 2008.

\bibitem{srinivasa2016survey}
C.~Srinivasa, S.~Ravanbakhsh, and B.~Frey.
\newblock {Survey Propagation beyond Constraint Satisfaction Problems}.
\newblock In {\em {Proceedings of {AISTATS}}}, pages 286--295, 2016.

\bibitem{stoop}
N.~Stoop, T.~Ott, and R.~Stoop.
\newblock Loopy belief propagation: Benefits and pitfalls on {Ising}-like
  systems.
\newblock In {\em Proceedings of {NOLTA}}. International Symposium on Nonlinear
  Theory and its Applications, 2006.

\bibitem{sutton2012improved}
C.~Sutton and A.~McCallum.
\newblock {Improved dynamic schedules for belief propagation}.
\newblock In {\em {Proceedings of {UAI}}}, 2007.

\bibitem{tatikonda2002loopy}
S.~C. Tatikonda and M.~I. Jordan.
\newblock {Loopy belief propagation and Gibbs measures}.
\newblock In {\em {Proceedings of {UAI}}}, 2002.

\bibitem{wainwright2003tree}
M.~J. Wainwright, T.~S. Jaakkola, and A.~S. Willsky.
\newblock Tree-reweighted belief propagation algorithms and approximate ml
  estimation by pseudo-moment matching.
\newblock In {\em Proceedings of AISTATS}, 2003.

\bibitem{watanabe}
Y.~Watanabe and K.~Fukumizu.
\newblock {Graph zeta function in the {Bethe} free energy and loopy belief
  propagation}.
\newblock In {\em {NIPS}}, pages 2017--2025, 2009.

\bibitem{weiss2000correctness}
Y.~Weiss.
\newblock {Correctness of local probability propagation in graphical models
  with loops}.
\newblock {\em Neural Comp.}, 12(1), 2000.

\bibitem{weller2015balanced}
A.~Weller.
\newblock Bethe and related pairwise entropy approximations.
\newblock In {\em Proceedings of {UAI}}, 2015.

\bibitem{weller2016uprooting}
A.~Weller.
\newblock Uprooting and rerooting graphical models.
\newblock In {\em International Conference on Machine Learning}, pages 21--29,
  2016.

\bibitem{weller2013bethe}
A.~Weller and T.~Jebara.
\newblock {Bethe bounds and approximating the global optimum}.
\newblock In {\em {Proceedings of {AISTATS}}}, 2013.

\bibitem{weller2013approximating}
A.~Weller and T.~Jebara.
\newblock {Approximating the {Bethe} partition function}.
\newblock In {\em {Proceedings of {UAI}}}, 2014.

\bibitem{weller2014clamping}
A.~Weller and T.~Jebara.
\newblock Clamping variables and approximate inference.
\newblock In {\em NIPS}, pages 909--917, 2014.

\bibitem{weller2014understanding}
A.~Weller, K.~Tang, T.~Jebara, and D.~Sontag.
\newblock Understanding the {Bethe} approximation: when and how can it go
  wrong?
\newblock In {\em Proceedings of {UAI}}, 2014.

\bibitem{welling2003approximate}
M.~Welling and Y.~Teh.
\newblock {Approximate inference in Boltzmann machines}.
\newblock {\em Artificial Intelligence}, 143(1):19--50, 2003.

\bibitem{welling2001belief}
M.~Welling and Y.~W. Teh.
\newblock {Belief optimization for binary networks: A stable alternative to
  loopy belief propagation}.
\newblock In {\em {Proceedings of {UAI}}}. Morgan Kaufmann Publishers Inc.,
  2001.

\bibitem{willsky2008loop}
A.~S. Willsky, E.~B. Sudderth, and M.~J. Wainwright.
\newblock {Loop series and {Bethe} variational bounds in attractive graphical
  models}.
\newblock In {\em {NIPS}}, 2008.

\bibitem{yedidia}
J.~S. Yedidia, W.~T. Freeman, and Y.~Weiss.
\newblock {Constructing free-energy approximations and generalized belief
  propagation algorithms}.
\newblock {\em IEEE Trans. on Information Theory}, 51(7):2282--2312, 2005.

\bibitem{cccp2003yuille}
A.~Yuille and A.~Rangarajan.
\newblock {The concave-convex procedure}.
\newblock {\em Neural Computation}, 15(4):915--936, 2003.

\bibitem{zdeborova2010generalization}
L.~Zdeborov{\'a} and F.~Krzakala.
\newblock Generalization of the cavity method for adiabatic evolution of gibbs
  states.
\newblock {\em Physical Review B}, 81(22), 2010.

\bibitem{zdeborova2016statistical}
L.~Zdeborov{\'a} and F.~Krzakala.
\newblock {Statistical physics of inference: Thresholds and algorithms}.
\newblock {\em Advances in Physics}, 65(5):453--552, 2016.

\end{thebibliography}
